\documentclass[sigconf]{acmart}
\usepackage{graphics}
\usepackage{graphicx}
\usepackage{color}
\usepackage{algorithm}
\usepackage{algorithmic}
\usepackage{booktabs}
\usepackage{subcaption}
\usepackage{makecell}
\usepackage{multirow}
\usepackage{adjustbox}
\usepackage[table]{xcolor}
\usepackage[dvipsnames]{xcolor}

\usepackage{amsmath, amssymb, amsthm}
\theoremstyle{plain}
\newtheorem{proposition}{Proposition}

\theoremstyle{remark}
\newtheorem{remark}{Remark}
\newcommand{\modela}{MHP}
\newcommand{\modelb}{A-MHP}
\AtBeginDocument{%
  }

\copyrightyear{2026}
\acmYear{2026}
\setcopyright{cc}
\setcctype{by}
\acmConference[WWW '26]{Proceedings of the ACM Web Conference 2026}{April 13--17, 2026}{Dubai, United Arab Emirates}
\acmBooktitle{Proceedings of the ACM Web Conference 2026 (WWW '26), April 13--17, 2026, Dubai, United Arab Emirates}
\acmPrice{}
\acmDOI{10.1145/3774904.3792583}
\acmISBN{979-8-4007-2307-0/2026/04}

\begin{document}

%\title{Time-variant Mamba Hawkes Process}
\title{Mamba Hawkes Process for Event Sequence Modeling}
% \iffalse
% \affiliation{%
% \fi

\author{Shan Dai}
\affiliation{%
  \institution{Shenzhen Research Institute of Big Data}
  \city{Shenzhen}
  \country{China}}
\email{shandai@sribd.cn}

\author{Yuyang Shen}
\affiliation{%
  \institution{The Chinese University of Hong Kong, Shenzhen}
  \city{Shenzhen}
  \country{China}}
\email{yuyangshen889@gmail.com}

\author{Yuyang Liang}
\affiliation{%
  \institution{The Chinese University of Hong Kong, Shenzhen}
  \city{Shenzhen}
  \country{China}}
\email{yuyangliang@link.cuhk.edu.cn}

\author{Chenhao Ma}
\authornote{Corresponding authors.}
\affiliation{%
  \institution{The Chinese University of Hong Kong, Shenzhen}
  \city{Shenzhen}
  \country{China}}
\email{machenhao@cuhk.edu.cn}

\author{Anningzhe Gao}
\authornotemark[1]
\affiliation{%
  \institution{Shenzhen Research Institute of Big Data}
  \city{Shenzhen}
  \country{China}}
\email{anningzhegao@gmail.com}

% \iffalse
% \affiliation{%

% \affiliation{%

% \affiliation{%

% \affiliation{%

% \affiliation{%

% \affiliation{%
% \fi

\renewcommand{\shortauthors}{Shan Dai, Yuyang Shen, Yuyang Liang, Chenhao Ma, \& Anningzhe Gao.}

\begin{abstract}

Modeling asynchronous event sequences is crucial in numerous real-world applications such as healthcare monitoring, financial transaction analysis, and so on. Traditional temporal point processes, including Hawkes Processes, often fail to capture complex dependencies due to their parametric limitations. While neural approaches like RNNs and Transformers have improved flexibility, they struggle with computational inefficiency, and attention saturation. In this paper, we introduce the Mamba Hawkes Process (MHP), the first framework to integrate selective state space model (Mamba) with temporal point processes. MHP leverages time-varying state transitions and input-dependent gating to efficiently encode event history and capture long-term dependencies with linear complexity. Importantly, we provide theoretical guarantees showing that MHP generalizes both classical multi-exponential Hawkes processes and exponential-decay gated RNNs, underscoring its expressive power and theoretical soundness. To address the inherent constraints of pure state space models in handling heterogeneous event interactions, we further develop Adaptive Mamba Hawkes Process (A-MHP) that incorporates two novel mechanisms: a Time-Scaling Mechanism that adaptively weights time intervals based on event type and history, and a Dual-Channel State Transition that adaptively processes event content and temporal dynamics for more refined state updates. Extensive experiments on synthetic and real-world datasets demonstrate that MHP and A-MHP consistently outperform state-of-the-art baselines in event prediction tasks, particularly in long-sequence scenarios. Our work establishes a scalable and theoretically grounded paradigm for event sequence modeling, with practical implications for predictive maintenance, anomaly detection, and dynamic system analysis. The code is available at https://github.com/Ethan-Shen-Individual-Lab/Mamba-Hawkes-Process.

\end{abstract}

\begin{CCSXML}
<ccs2012>
   <concept>
       <concept_id>10010147.10010257</concept_id>
       <concept_desc>Computing methodologies~Machine learning</concept_desc>
       <concept_significance>500</concept_significance>
       </concept>
 </ccs2012>
\end{CCSXML}

\ccsdesc[500]{Computing methodologies~Machine learning}


\keywords{Asynchronous event, Hawkes Process, State Space Model, Time-varying State, Long-term dependency}


\maketitle

\section{Introduction}

Large volumes of irregular and asynchronous event sequences are generated by both human activities and natural phenomena. Examples include user interactions on social media platforms \citep{zhang2023learning, fang2023group}, transaction histories in financial markets \citep{bacry2015hawkes, hawkes2018hawkes}, electronic health records \citep{wang2018supervised}, and earthquake occurrences with aftershocks in geophysics \citep{ogata1998space}. Unlike traditional sequential data such as time series, these asynchronous event sequences are characterized by irregular intervals between events, where the timing of events is as essential as their sequence order in describing underlying dynamics.

Temporal point processes (TPPs), defined by the intensity functions, are widely used for modeling asynchronous event sequences \citep{cox1980point,brillinger2002point}. Among TPPs, Hawkes Processes are particularly noteworthy for their non-negative intensity functions that capture the triggering effects of previous events. However, traditional Hawkes Processes have significant limitations. They do not account for mutual inhibition between events, a crucial factor in many real-world scenarios, and they lack robust nonlinear fitting capabilities, thereby limiting their expressive power. To address these limitations, researchers have introduced likelihood-free methods \citep{xiao2017wasserstein,li2018learning} and also non-parametric approaches like kernel methods and splines \citep{zhou2013learning}.

\begin{figure}[thb]
    \centering
    \includegraphics[width=0.8\linewidth]{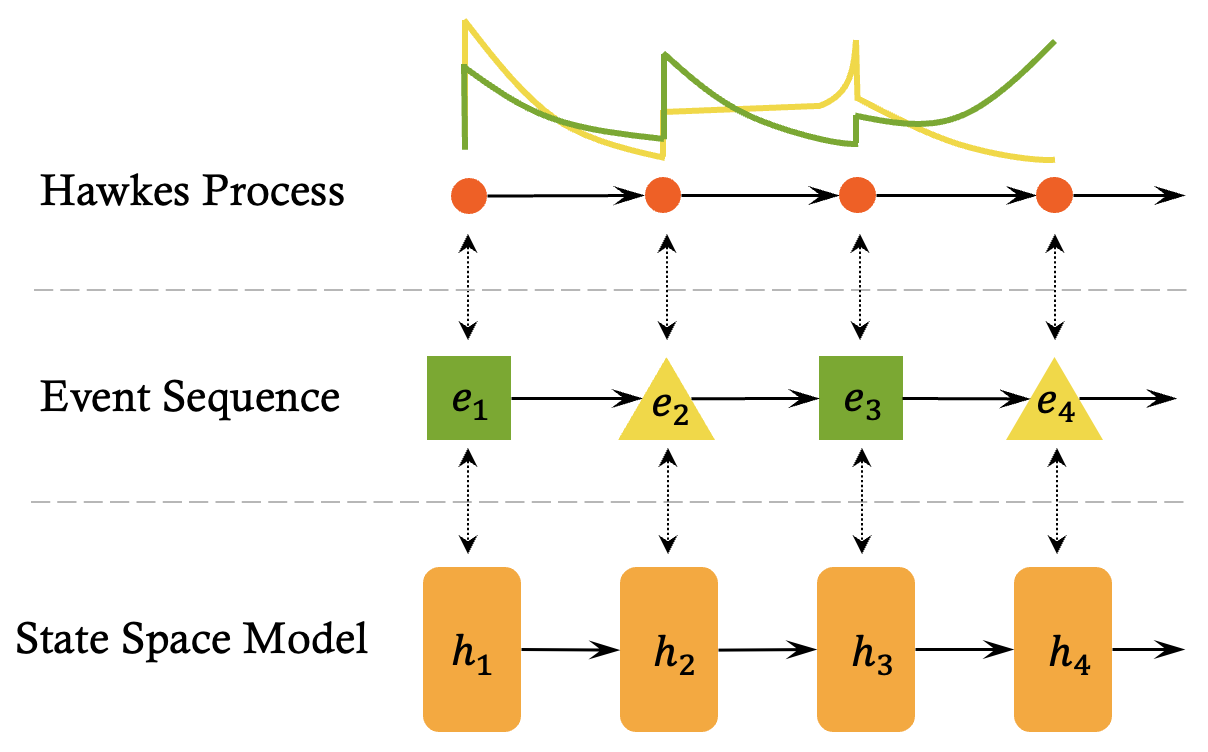}
    \caption{SSM is naturally aligned with the Hawkes Process}
    \label{fig:intro}
    \vspace{-0.5cm}
\end{figure}

% The advancement of neural networks and deep learning has further revolutionized sequence modeling. \cite{du2016recurrent} introduced the use of Recurrent Neural Networks (RNNs) to model events propagation, leading to the development of RNN-based Hawkes process models. This approach allows for neurally self-modulating multivariate point processes by not requiring historical contributions to be additive, and it enables the modeling of complex memory effects, such as delays.

% Despite these advances, RNN-based models have notable drawbacks. Even with mechanisms such as Long Short-Term Memory (LSTM) \citep{hochreiter1997long} and Gated Recurrent Units (GRUs) \citep{chung2014empirical}, RNNs notoriously struggle to learn long-term memories, primarily due to vanishing and exploding gradients \citep{pascanu2013difficulty}. To address these challenges, \cite{zuo2020transformer} proposed the Transformer Hawkes Process (THP) model, which leverages a transformer architecture devoid of RNNs or CNNs and achieves state-of-the-art performance. %Nonetheless, transformer models can be further refined as they simply stack encoder layers to learn sequence data, potentially neglecting the recursive bias learning inherent in RNNs. 
% Nevertheless, attention-based transformers also encounter limitations in modeling long input sequences,
% especially when dependencies extend far beyond the attention window. This in-context constraint is especially significant in long-sequence prediction tasks, as demonstrated by \cite{gao2024rothp}.
The advancement of neural networks and deep learning has revolutionized sequence modeling. \cite{du2016recurrent} introduced RNN-based Hawkes process models, enabling neurally self-modulating point processes and capturing complex memory effects. However, RNNs, even with LSTM \citep{hochreiter1997long} and GRU \citep{chung2014empirical}, struggle with long-term dependencies due to vanishing and exploding gradients \citep{pascanu2013difficulty}. To overcome this, \cite{zuo2020transformer} proposed the Transformer Hawkes Process (THP), which avoids RNN/CNN structures and achieves better performance. Yet, attention-based transformers also face challenges in modeling very long sequences, as their finite attention windows limit long-range dependency capture \cite{gao2024rothp}.

% Recently, structured state space sequence models (SSMs) \citep{gu2021efficiently} have emerged as a promising class of architectures for sequence modeling. The Mamba model \citep{gu2023mamba}, a selective state space model, addresses data-dependent context compression in sequence modeling. Unlike attention mechanisms, Mamba utilizes state space constructs to encode context using hidden states during recurrent scans. The selection mechanism determines which parts of the input influence the hidden states, thereby guiding subsequent embedding updates. The dynamics of temporal point processes are also described by a continuous conditional intensity function. Mamba shares the property of continuous dynamic mechanism modeling, which shows the potential to deal with the complexity inherent in TPPs.

% Recently, structured state space sequence models (SSMs) \citep{gu2021efficiently} have emerged as a promising class of architectures for sequence modeling. The Mamba model \citep{gu2023mamba}, a selective SSM, introduces data-dependent context compression by updating hidden states through a selection mechanism that determines which inputs influence subsequent embeddings. Similarly, temporal point processes such as Hawkes Processes and so on describe event dynamics via a continuous conditional intensity function. This shared focus on continuous dynamic modeling highlights the natural alignment between Mamba and Hawkes Processes, suggesting strong potential for handling the complexities of TPPs (Figure \ref{fig:intro}). 

Recently, structured state space sequence models (SSMs) \citep{gu2021efficiently} have emerged as a powerful paradigm for modeling long sequences with continuous dynamics. The Mamba model \citep{gu2023mamba}, a selective SSM, introduces data-dependent context compression by updating hidden states through a continuous transition kernel that controls how past information decays or persists over time. This mechanism closely parallels temporal point processes (TPPs), where each event modulates a continuous conditional intensity function. In both frameworks, the latent dynamics evolve between events and are updated upon event occurrences. Consequently, selective SSMs like Mamba offer a natural neural parameterization for temporal point processes (Figure~\ref{fig:intro}).

However, adapting Mamba to model Hawkes Processes introduces several theoretical and practical challenges. 
% First, Mamba inherently emphasizes sequential dependency through its time-varying hidden states, but aligning this sequential bias with the event-driven, discontinuous nature of TPPs requires redefining state transitions to dynamically adapt to irregular event arrivals. 
%
First, Mamba and other SSMs are typically designed for regularly sampled sequences, whereas Hawkes Processes operate on irregular, event-driven timelines, thereby requiring the state transitions of SSMs to be redefined to adapt to non-uniform event arrivals.
Second, Mamba’s linear state space formulation, while computationally efficient, limits its ability to represent nonlinear interactions common in real-world event dynamics. Third, while Mamba excels at capturing long-range dependencies, its selective scanning mechanism may overlook sparse but critical long-term excitatory or inhibitory effects in TPPs, such as delayed reactions in social contagion \citep{ding2021quantifying}.
To address these challenges, we propose a novel Mamba Hawkes Process (\modela) model that integrates Mamba’s selective state space framework with Hawkes Processes.
Beyond this intuitive alignment, we provide theoretical guarantees that substantiate the connection between Mamba and temporal point processes. Specifically, we prove that the proposed MHP can be degraded into a multi-exponential Hawkes process as well as an exponential-decay gated RNN in Section \ref{sec:property}.

While \modela~ establishes a principled foundation, its  state transition nature and fixed temporal scaling limit its expressive power in capturing heterogeneous event interactions in sequences. To overcome these limitations, we further propose the Adaptive Mamba Hawkes Process (\modelb), an extended framework that enhances the expressiveness of \modela~ by introducing adaptive temporal processing and dual embedding transitions to better capture heterogeneous event dynamics. 

{\bf Contributions.} Building on these innovations, the main contributions of this work are summarized as follows:
\begin{itemize}
	\item \textbf{Constructive Integration of Mamba into Temporal Point Processes.} To the best of our knowledge, this is the inaugural work to adapt the Mamba state space model for temporal point processes, enabling efficient event history encoding and improved long-term dependency modeling beyond RNNs and Transformers. Importantly, we provide theoretical foundations proving that MHP can be reduced to multi-exponential Hawkes Processes and exponential-decay gated RNNs under specific configurations. This alignment not only validates the design of MHP but also highlights its generality and expressive capacity in capturing complex temporal dynamics.
    % First Integration of Mamba Architecture into Temporal Point Processes. To the best of our knowledge, this is the inaugural work to adapt the Mamba state space model, a breakthrough in long-sequence modeling, for temporal point processes. By designing time-variant state transition parameters and gating mechanisms, MHP dynamically adjusts its hidden states to encode event history, effectively addressing the limitations of RNNs and Transformers in computational inefficiency and inadequate handling of long-term dependencies. This innovation enables linear-time computation while capturing intricate dependencies in sequences spanning long events.
	% \item \textbf{A Unified and Extensible Framework Enabling Hybrid Modeling.} Benefitting from the flexible extensibility, MHP is readily integrated with other models. We propose the \modelb, 
    \item \textbf{Adaptive Temporal Modeling with Dual Innovations.} To address the inherent constraints of a state space model in handling heterogeneous event interactions, we develop \modelb, which introduces two key mechanisms- \textit{Time-Scaling} for adaptive temporal processing across event types, and \textit{Dual-Channel State Transition} for content-dependent and time-dependent embedding computation, addressing limitations of \modela~ and enhancing event prediction accuracy.
    
	\item \textbf{Comprehensive Experiment Validation and Practical Impact.} We conducted extensive experiments on both synthetic and real-world datasets to assess the effectiveness of \modela~ and \modelb~, which consistently outperform state-of-the-art baselines in both event type and timing and enable more accurate event prediction in practical applications.
    % We conducted extensive experiments on synthetic and real-world datasets including high-frequency financial transactions, electronic health records, and social network activity to evaluate the effectiveness of our model, with a focus on performance across varying sequence lengths. MHP and TV-MHP outperform state-of-the-art baselines in various metrics, particularly excelling in long sequences where existing models degrade. The results demonstrate that MHP and its extension consistently outperform competing models in predicting both event type and timing, particularly in tasks involving longer sequences. These advancements directly support applications such as event prediction in cyber-physical systems and early detection in intelligent healthcare.
	
\end{itemize}

\section{Related Work}
\label{sec:2}

\subsection{Neural Hawkes Process}
Hawkes processes are widely used for temporal prediction in various fields. To enhance their performance, many deep learning approaches have been applied.
The Recurrent Marked Temporal Point Process (RMTPP) \cite{du2016recurrent} employs RNNs to model the influence of event history on intensity functions. RMTPP inspired subsequent models that use dual RNNs, one for background intensity and another for the impact of historical events, enabling end-to-end sequence modeling \cite{xiao2017modeling}. Continuous-time LSTM models \cite{mei2017neural} extend RMTPP by capturing self-modulating effects, addressing both excitation and inhibition from previous events.

With the development of self-attention mechanisms, self-attention-based neural Hawkes processes were proposed. SAHP \cite{zhang2020self} utilized self-attention to enhance these processes, while THP \cite{zuo2020transformer} used the transformer encoder to convert sequence data into continuous conditional intensity functions. UTHP \citep{zhang2021universal} incorporated RNNs and CNNs to address issues in THP, such as parallel processing and recursive learning. TAA-THP \citep{zhang2022temporal} improved attention structures by incorporating temporal encoding. Lastly, \cite{yu2022transformer} proposed Hawkesformer, linking hierarchical attention mechanisms to Hawkes processes for information cascade prediction.

\subsection{State Space Model}

State space models (SSMs) were initially developed as mathematical tools to describe dynamic systems in modern control theory. With the introduction of HiPPO \citep{gu2020hippo} initialization, the Linear State-Space Layer (LSSL) \citep{gu2021combining} demonstrated the capability to handle long-range dependencies in sequential data. However, LSSL's computational and memory overhead make it impractical for large-scale applications.  Structured  State Space models (S4) \citep{gu2021efficiently} address these issues by using reparameterization to enhance computational efficiency, providing an effective alternative to traditional attention mechanisms.

Several recent variants of S4 have been proposed to achieve linear time attention, including H3 \citep{fu2022hungry}, Gated State Space \citep{mehta2022long}, Hyena \citep{nguyen2024hyenadna}, and RWKV \citep{peng2023rwkv}. Mamba \citep{gu2023mamba} introduces a data-dependent selection mechanism to S4, improving the capture of long-range context as sequence length increases. Mamba not only achieves linear time efficiency in long-sequence modeling but also outperforms Transformers across various benchmarks. Recently, Jamba \citep{lieber2024jamba} has been introduced as a novel hybrid model that combines Transformer and Mamba layers in a mixture-of-experts (MoE) architecture. Jamba interleaves blocks of Transformer and Mamba layers, harnessing the strengths of both model families.
\section{Preliminaries}
\label{sec:3}

\subsection{Hawkes Processes}
A TPP \citep{daley2003introduction,daley2007introduction} is a stochastic process that defines a probability distribution over event sequences. In this process, the number of points (events) $K$ and their locations (arrival times) $t_{i}$ are random. The realization of a TPP can be represented as a sequence of discrete events with $\{t_{i} \} \in \mathbb{R}^{+}$ and $i \in \mathbb{Z}^{+}$ abstracted as points on a timeline. We can represent a TPP realization by a counting measure $N(t)=\sum_{i}^{n} \mathbb{I}(t_{i}<t)$, for $t \in [0,T]$. The conditional intensity characterizing a TPP can be interpreted as the expected number of events per unit of time and is defined as:
\begin{equation}\label{eq1}
	%\begin{align*}
	\lambda(t)=\lim\limits_{\Delta t \downarrow 0} \frac{\mathbb{E}[N(t+\Delta t)-N(t) | \mathcal{H}_{t}]}{\Delta t},
	%\end{align*}
\end{equation}
where $\mathcal{H}_{t}=\{ t_{i}: t_{i}<t \}$ is the event history until time $t$, which acts as a filtration to the process. {Given $\lambda(t)$, the probability that no new event occurs up to time $t$, after the last event occurblue at $t_n$, can be expressed as $\exp \left( - \int_{t_n}^{t} \lambda(\tau) \, d\tau \right)$. Then, the conditional density function can be formulated by
	\begin{equation}\label{eq:density}
		f(t) = \lambda(t) \exp \left( - \int_{t_n}^{t} \lambda(\tau) \, d\tau \right).
\end{equation}}

The Hawkes process \citep{hawkes1971spectra,laub2015hawkes} is a typical temporal point process, and it models past events and predicts the timestamp of the following event by its conditional intensity function, which is defined as: 
\begin{equation}\label{eq2}
	\lambda (t)= \mu + \sum\limits_{j: t_{i}<t} \psi (t-t_{i}),  
\end{equation}
where $\mu \geq 0$, named base intensity, is an exogenous component of the intensity function independent of the history, while $\psi(t)>0$ is an endogenous component dependent on the history. Besides, $\psi(t)$ is a triggering kernel containing the peer influence of past events. Due to the self-exciting charities, the Hawkes process has received wide attention and applications in event sequence modeling.

\subsection{Mamba}
The structublue state space sequence models (S4) \citep{gu2021efficiently} is defined by the equation \eqref{eq:ssm}, which maps a 1-dimensional function or sequence $x(t) \in \mathbb{R} \mapsto y(t) \in \mathbb{R}$ through an implicit latent state $h(t)\in \mathbb{R}^{N}$:
\begin{equation}
	\begin{aligned}
		h^{\prime}(t) &= \mathbf{A}h(t) + \mathbf{B}x(t),  \\
		y(t) &= \mathbf{C}h(t),
		\label{eq:ssm}
	\end{aligned}
\end{equation}
where $\mathbf{A} \in \mathbb{R}^{N \times N}, \mathbf{B} \in \mathbb{R}^{N \times 1}, \mathbf{C} \in \mathbb{R}^{1 \times N}$ are parameters of neural networks in deep learning. To deal with the discrete input sequence $\boldsymbol{x} = (x_0, x_1, ...)  \in \mathbb{R}^{L}$, S4 discretizes these parameters in equation \eqref{eq:ssm} using a step size $\Delta$, where the continuous parameters $\mathbf{A}, \mathbf{B}$ are converted into discrete parameters $\overline{\mathbf{A}}=f_{A}(\Delta,\mathbf{A}), \overline{\mathbf{B}}=f_{B}(\Delta,\mathbf{B})$, where the pair $(f_{A},f_{B})$ is called a discretization rule \citep{gu2021efficiently}. {As a result, a recurrent formula is derived to iteratively compute the state and output of the system for discrete time steps}:

\begin{equation}
	\begin{aligned}
		h_t &= \mathbf{\overline{A}} h_{t-1} + \mathbf{\overline{B}} x_t, \\
		y_t &= \mathbf{C} h_t.
		\label{eq:ssmd}
	\end{aligned}
\end{equation}

Various rules can be used such as the zero-order hold (ZOH) defined as follows:

\begin{equation}
	\begin{aligned}
		\overline{\mathbf{A}} &= \exp(\Delta\mathbf{A}), \\
		\overline{\mathbf{B}} &= (\Delta\mathbf{A})^{-1}(\exp(\Delta\mathbf{A}) - \mathbf{I}) \cdot \Delta\mathbf{B}.
	\end{aligned}
\end{equation}

{In addition to recurrent mode in equation \eqref{eq:ssmd}, the model can also be computed in convolution mode, as expanded below}:
\begin{equation}
	\begin{aligned}
		\overline{\mathbf{K}} &= (\mathbf{C}\overline{\mathbf{B}}, \mathbf{C}\overline{\mathbf{AB}}, \ldots, \mathbf{C}\overline{\mathbf{A}}^{L-1}\overline{\mathbf{B}}), \\
		\boldsymbol{y} &= \boldsymbol{x} * \overline{\mathbf{K}},
		\label{eq:ssm_c}
	\end{aligned}
\end{equation}
where $L$ denotes the length of the input sequence $\boldsymbol{x}$ and $\overline{\mathbf{K}} \in \mathbb{R}^{L}$ is the convolution kernel.

By introducing a data-dependent selection mechanism to S4, the Mamba models \citep{gu2023mamba} utilize selective SSMs with a time-variant machanism, where the discrete matrices $\bar{A},\bar{B},$ and $C$ of each channel are modified over $L$ time steps depending on the input sequence. Unlike traditional state-space layers, which operate individually on each channel, selective state-space layers compute the SSM matrices $\bar{A}_i, \bar{B}_i, C_i$ for all $i \leq L$ based on all the channels, and then apply the time-variant recurrent rule individually for each channel. 

Thus, we denote the entire input sequence by $\hat{x} := (\hat{x}_1, \cdots, \hat{x}_L) \in \mathbb{R}^{L \times D}$ where $\hat{x}_i \in \mathbb{R}^{D}$. The discrete matrices $\mathbf{B}_i, \mathbf{C}_i,$ and $\Delta_i$ are defined as follows:
\begin{equation} \label{eq:TimeVariantMatrices1}
	\mathbf{B}_i = S_B (\hat{x}_i), \quad \mathbf{C}_i = S_C (\hat{x}_i), \quad \Delta_i = \text{Softplus}(S_{\Delta}(\hat{x}_i)).
\end{equation}
By defining
\begin{equation} \label{eq:discretization}
	\begin{aligned}
		f_A(\Delta_i, \mathbf{A}) &= \exp (\Delta_i \mathbf{A}), \\ 
		f_B(\Delta_i, \mathbf{A}, \mathbf{B}_i) &= (\Delta_i {\mathbf{A}})^{-1}(\exp(\Delta_i {\mathbf{A}}) - \mathbf{I}) \cdot\Delta_i \mathbf{B}_i,
	\end{aligned} 
\end{equation}
where $f_A, f_B$ represents the discretization rule, $S_B, S_C, S_{\Delta}$ are linear projection layers, and the Softplus is an elementwise function that is a smooth approximation of ReLU. Then we get:
\begin{equation}\label{eq:TimeVariantMatrices2}
	\overline{\mathbf{A}}_i = f_A(\Delta_i, \mathbf{A}), \quad \overline{\mathbf{B}}_i = f_B(\Delta_i, \mathbf{A}, \mathbf{B}_i).
\end{equation}

\section{Mamba Hawkes Process}\label{process}

The primary challenge in modeling event sequence data revolves around several aspects. Firstly, the discontinuous nature of TPPs requires redefining state transitions to dynamically adapt to irregular event arrivals, which involves designing input-dependent gating mechanisms to reset or propagate hidden states only at event timestamps rather than fixed intervals, and poses a non-trivial adaptation problem in standard SSMs.
Secondly, there is the question of how to efficiently handle extended event sequences while effectively capturing the complex and nonlinear evolving dynamics of the intensity function that drives the sequence. Additionally, it is crucial to capture long-range event transition dependencies within the sequence, particularly considering the self-exciting properties inherent to a large class of point processes that may exhibit interactions between events located far apart in the temporal domain. To address these challenges, we propose the Mamba Hawkes Process (\modela) model.

Denote the sequence $\mathcal{S} = \{(t_j, k_j) \}_{j=1}^L$ as the temporal event sequence of $L$ events, where each events has type $k_j\in\{1,2,...,K\}$ with a total number of $K$ types. Each pair $(t_j, k_j)$ corresponds to an event of type $k_j$ occurring at time $t_j$. Different from \eqref{eq:TimeVariantMatrices1}, let $\Delta_j = t_j - t_{j-1}$ represent the temporal differences, with $\Delta_1 = t_1$ for convention, hence the temporal sequence can be represented by
$\mathbf{\Delta}=(\Delta_1,\Delta_2,...,\Delta_L)$.

%\subsection{Time-variant Recurrent State}\label{recurrent_state}

\subsection{Model Formulation}\label{recurrent_state}

% This section demonstrates the basic design of MHP. 
Let $W^e$ be the event embedding matrix with dimensions $D\times K$, where the $k$-th column of $W^e$ is a $D$-dimensional embedding for event type $k$. For any event of type $k_j$, let $\mathbf{k}_j$ be its one-hot encoding, then the event embedding at corresponding timestamp $t_j$ is calculated as $x_{t_j} = \mathbf{k}_i(W^e)^T$. Thus, for the event sequence $\mathcal{S}$, the embedding is computed by
\begin{equation}
	(x_{t_1},x_{t_2},...,x_{t_L})=(\mathbf{k}_1,\mathbf{k}_2,...,\mathbf{k}_L)(W^e)^T.
\end{equation}

The key idea is to enable the Mamba model to handle the temporal dependency over both the event types and the timestamps from past events. Equivalent to equation \eqref{eq:ssmd}, we have discrete recurrence: 
\begin{equation}
	h(t+\Delta) = \overline{\mathbf{A}}h(t) + \overline{\mathbf{B}}x(t),
	\label{equ:des}
\end{equation}
where the hidden state at $(t+\Delta)$ is built on previous state at $t$. In the Mamba architecture, the parameters $(\Delta,\mathbf{B},\mathbf{C})$ are time-dependent and are obtained by linear projection from the input, thus resulting in a time-variant SSM. To model the event sequence $\mathcal{S}$, we adopt the time-variant version of equation \eqref{equ:des} and construct Mamba Hawkes Process as Algorithm~\ref{alg:architecture}. For the event of type $k_j$ occurring at time $t_j$, the event embedding $x_{t_j}$ is fed as input into the state space model, and temporal difference between this event and its previous event $\Delta_j = t_j-t_{j-1}$ is assigned as the value of $\Delta$.  

\begin{algorithm}[t!]
	\caption{The architecture of Mamba Hawkes Process}
	\label{alg:architecture}
	{\bf Input:}
	The temporal sequence $\{(t_1, k_1),(t_2,k_2),...,(t_L,k_L)\}$
	{\bf Output:}
	The hidden state $\mathbf{h}$
	\begin{algorithmic}[1]
		\STATE $A:(L,D)\gets \text{Parameter}$ // $N\times N$ matrix
		\STATE $B:({B},L,D) \gets \text{Linear}_B(x_{t_j})$ // 
		\STATE $C:({B},L,D) \gets \text{Linear}_C(x_{t_j})$ // $B$ and $C$ are time-dependent
		\STATE $\Delta\gets t_{j}-t_{j - 1}$ // $\Delta$ records the temporal information
		\STATE $\Bar{A}, \Bar{B}\gets \text{Discretize} (A,B,\Delta)$
		\STATE $\mathbf{h}\gets \textbf{SSM}(\Bar{A},\Bar{B},C,\Delta)(x)$
		
		\RETURN $\mathbf{h}$
	\end{algorithmic}
\end{algorithm}

In particular, We define
$$\mathbf{B}(t_j)=\text{Linear}_\mathbf{B}(x_{t_j}),\mathbf{C}(t_j)=\text{Linear}_\mathbf{C}(x_{t_j}),$$
which are obtained by a linear transformation of the vector $x_{t_j}$. We have the transition formulas for the Mamba Hawkes process:
\begin{equation}
	\begin{aligned}
		h_{t_j} &= \overline{\mathbf{A}}(t_j)h_{t_{j-1}} + \overline{\mathbf{B}}(t_j)x_{t_j}, \\
		y_{t_j} &= \mathbf{C}(t_j)h_{t_j},
		\label{eq:mamba_d}
	\end{aligned}
\end{equation}
where
\begin{equation}
	\begin{aligned}\label{mhp}
		\overline{\mathbf{A}}(t_j) &= \exp(\Delta_j\mathbf{A}), \\
		\overline{\mathbf{B}}(t_j) &= (\Delta_j\mathbf{A})^{-1}(\exp(\Delta_j\mathbf{A}) - \mathbf{I}) \cdot \Delta_j\mathbf{B}(t_j),
	\end{aligned}
\end{equation}
are the temporal-dependent coefficients. As shown in equation \eqref{eq:mamba_d}, the hidden state $h_{t_{j-1}}$ represents the memory of the influence from the timings and the types of past events up to timestamp $t_{j-1}$. The time-variant SSM updates $h_{t_{j-1}}$ to $h_{t_j}$ by taking into account the effect of the current event $(t_j, k_j)$ in a selective way. Specifically, the model modulate the recurrent dynamics based on temporal difference $\Delta_j$ and current input $x_{t_{j-1}}$, determining how much to forget or persist the current state $h_{t_{j-1}}$ and whether to focus or ignore the current input.

{Once the hidden representation of the event sequence is generated, we can further interpolate the continuous intensity function. Given $h_{t_j}$, the conditional} intensity function of Neural Hawkes process is given by
\begin{equation}
	\lambda(t) = \sum_{k=1}^K\lambda_k(t),
\end{equation}
where $\lambda_k$ is the {conditional} intensity function of the $k$-th event, and 
\begin{equation}
\begin{aligned}\label{eq:intensity}
	\lambda_k&=f_k(\alpha_k(t-t_j) + \mathbf{w}_k^T\mathbf{h}(t_j) + b_k),\\
    f_k(x)&=\beta_k \log \left(1+\exp \left(x / \beta_k\right)\right),
\end{aligned}
\end{equation}
where $t$ is defined on interval $t \in\left[t_j, t_{j+1}\right)$, and $f_k(x)$ is the Softplus function. 
The Softplus function is chosen for the intensity function in Hawkes process models for two primary reasons. First, it ensures that the intensity remains non-negative. Second, it is smooth and differentiable everywhere, which facilitates training with gradient-based optimization.
The first term $\alpha_k(t-t_j)$ in equation \eqref{eq:intensity} models the temporal decaying effect of past events of type $k$ in Hawkes process. Specifically, $t-t_j$ represents the time elapsed since timestamp $t_j$, and $\alpha_k$ controls how quickly the influence of past events of type $k$ diminishes over time.
The second term $\mathbf{w}_k^T\mathbf{h}(t_j)$ represents the cumulative influence of all past events of type $k$ up until timestamp $t$, reflecting the self-modulating feature of the Hawkes process.
The third term $b_k$ indicates the baseline level of intensity for the event type $k$, independent of the history and elapsed time.

\subsection{Model Training}

{Given the conditional intensity function, the log-likelihood $\mathcal{\ell}$ of the event sequence $\mathcal{S}^i$ can be derived based on equation \eqref{eq:density}, and is expressed as}
\begin{equation}
	\mathcal{\ell}(\mathcal{S}^i) =  \sum_{i=1}^L \log \lambda_{}\left(t_i\right)-\int_{t_{1}}^{t_{L}} \lambda(t) d t.
\end{equation}
Based on this, we can further obtain the log-likelihood for a collection of sequences $\mathcal{S}=\mathcal{\{S}^i\}_{i=1}^N$, where $\mathcal{S}^i = \left( (t_j^i, k_j^i) \right)_{j=1}^{L_i}$, which is given by
	\begin{equation}
		\mathcal{\ell(\mathcal{S})} = \sum_{i=1}^N \mathcal{\ell}(\mathcal{S}^i).
	\end{equation}
The model parameters are learned by maximizing the log-likelihood across all sequences $\mathcal{\ell(\mathcal{S})}$. This optimization problem can be efficiently solved by stochastic gradient type algorithms like ADAM \cite{kingma2014adam},
except one challenge in computing $\int_{t_{1}}^{t_{L}} \lambda(t) d t$, the log-likelihood for the non-event part. Since it involves the softplus function, there is no closed-form computation for this integral, and we adopt a similar approximation computing method \cite{zuo2020transformer} for the non-event log-likelihood by using Monte Carlo integration.

To facilitate the prediction of the next event’s type and timestamp, we train two linear layers that operate on the hidden state representation \( \mathbf{h}(t_j) \in \mathbb{R}^D \), which encodes historical information up to the \( j \)-th event. Specifically, the layer for predicting next event type is denoted as \( \mathbf{P}^e \in \mathbb{R}^{K \times D} \), and the layer for predicting next event timestamp is denoted as \( \mathbf{P}^t \in \mathbb{R}^{1 \times D} \). Building on this, the next event type \( \hat{k}_{j+1} \) and the timestamp \( \hat{t}_{j+1} \) can be predicted as follows: 
\begin{equation}
\begin{aligned}
	\hat{k}_{j+1} &=\text{argmax}(\text{Softmax}(\mathbf{P}^e\mathbf{h}(t_j)),\\
	\hat{t}_{j+1} &= \mathbf{P}^t\mathbf{h}(t_j).
\end{aligned}
\end{equation}
Given the previously defined collection of sequences $\mathbf{S}$, we define two primary loss function to learn \( \mathbf{P}^e \) and \( \mathbf{P}^t \), specified by
\begin{equation}
\begin{aligned}
	\mathcal{L}_{event}(\mathcal{S}) &= \sum_{j=1}^{n-1}-\log(\text{Softmax}(P^e\mathbf{h}(t_j))_{k_{j+1}}),\\
	\mathcal{L}_{time}(\mathcal{S}) &= \sum_{j = 1}^{n-1}((t_{j+1} - t_j) - (\hat{t}_{j+1} -\hat{t}_j))^2,
\end{aligned}
\end{equation}
where ${t}_j$ is the true timestamp of event $j$. $\mathcal{L}_{event}$ is the cross-entropy loss that measures the accuracy of the event type prediction, and $\mathcal{L}_{time}$ is the MSE loss that measures the accuracy of the time prediction. {Finally}, the joint training loss can then be defined as
\begin{equation}\label{eq15}
	\mathcal{L}(\mathcal{S}) = -\mathcal{\ell}(\mathcal{S}) + \beta\mathcal{L}_{event}(\mathcal{S}) + \gamma\mathcal{L}_{time}(\mathcal{S}),
\end{equation}
where $\beta,\gamma$ are hyperparameters to control the range of event and time losses.

\subsection{Model Properties}\label{sec:property}
	\subsubsection{\textbf{Relations to Hawkes processes}}

	We show MHP contains classical multi-exponential Hawkes processes as a special case.
	
	\begin{proposition}[Recovery of multi-exponential Hawkes]
		\label{prop:multi-exp}
		Consider a $R$-dimensional state with $A=\mathrm{diag}(-\omega_1,\dots,-\omega_R)$, $\omega_r>0$ and   $B\in\mathbb{R}^{R\times K}$ as a learnable matrix in \eqref{mhp}. Let $x(t)$  be $ \sum_{j=1}^{n} \delta(t-t_j)\,\mathbf{k}_j $ where $\mathbf{k}_j\in\mathbb{R}^K$ is the one-hot event type vector encoding type ${k}_j$, and  the  intensity function be linear and nonnegative:
		$
		\lambda_k(t)= \mu_k + \sum_{r=1}^R c_{kr} h_r(t),
		$
		with $\mu_k\ge 0$ and $c_{kr}\ge 0$. Then $\lambda_k(t)$ equals a Hawkes intensity with $K$-to-$K$ triggering kernels
		$
		\psi_{k k'}(u) = \sum_{r=1}^R c_{k r} B_{r k'} e^{-\omega_r u}, \\\quad u>0.
		$
	\end{proposition}
Proposition 1 shows the reduction of \modela~ to classical Hawkes Process by selecting a linear,
nonnegative readout; with Softplus, one obtains a flexible nonlinearity around the linear regime. 

        \subsubsection{\textbf{Relations to neural point processes}}
    Now we show that \modela~ can be degraded to an exponential-decay gated RNN.
    	\begin{proposition}[Degeneration to an exponential-decay gated RNN]
		\label{prop:edrnn}
			\label{prop:gated-rnn}
		Let $A=-\omega I$ with $\omega>0$,  scalar input held piecewise constant on $[t_{i-1},t_i)$ with value $x_i$, and constant $\mathbf{B}(t_j)= \omega I$ for different $t_j$. Under ZOH discretization at event times, the MHP update reduces to
		\begin{equation}
		h_i = e^{-\omega \Delta_i} h_{i-1} + \big(1 - e^{-\omega \Delta_i}\big)\, x_i,
		\end{equation}
		i.e., a time-aware exponentially decayed gated RNN with gate $g_i = e^{-\omega \Delta_i}\in(0,1)$.
	\end{proposition}

Proposition 2 shows that \modela generalizes RMTPP’s intensity head while adopting a different, event-driven, selective SSM recurrence
with continuous-time consistency and linear-time scanning. Detailed proofs of Proposition \ref{prop:multi-exp} and \ref{prop:edrnn} can be found in the appendix.

\section{Adaptive Mamba Hawkes Process}
\subsection{Time-Scaling Mechanism}\label{Time-Scaling}

\iffalse
Based on the time-variant recurrent state designed at \eqref{recurrent_state}, we further address two progressive structure improvements. The first is the Time-Scaling mechanism, where we replace the physical time intervals $\Delta_j$ with scaled time intervals $\Delta_j'$. Then comes Information Embedding, where we concatenate time and event type information to reformulate $\mathbf{B}(t_j)$. The proposed enhancements specifically targeted two critical aspects of the Mamba architecture: temporal discretization and the selection mechanism, which largely influence the model's efficiency.
\fi

While the foundational \modela~ framework provides significant advantages, our analysis revealed an important limitation: conventional temporal discretization assumes uniform decay rates across event types. This contradicts empirical evidence showing varying temporal dynamics—financial transactions exhibit persistent liquidity impacts while social media interactions demonstrate rapid decay. To address this, we developed the time-scaling mechanism, which introduces adaptive temporal processing.

Time-Scaling design is based on the recognition that different time intervals within a temporal point process should be weighted distinctly. For example, the weights should be influenced by the type of events, which is natural in real temporal processes such as earthquakes and financial orderings, where high-magnitude earthquakes or large-volume orders should be given special attention, as they exert considerably larger impact on the fitting of point processes. We also propose that these weights be determined not only by the type of event at the current time $t_j$, but also by the historical experience observed by the model, which dictates whether the weights should be amplified or attenuated. Therefore, we designed the Time-Scaling mechanism as follows:
\begin{equation}
\begin{aligned}
\Delta_j' &= \Delta_j\cdot\mathbf{s}(\mathbf{L}_j),\\
\mathbf{s}(\mathbf{L}_j) &= \text{Softplus}(W_{l}^Tl_j+b_{l}),\\
l_j &= \mathbf{GRU}(\mathbf{x}_{t_{j}},l_{j-1}),\label{time_scaling}
\end{aligned}
\end{equation}
where $\Delta_j'$ is the scaled time interval that is actually used in the latter training process. $\mathbf{s}(\mathbf{L}_j)$ is the scaling factor, which is calculated based on the hidden layer $L_j$, which stores the chosen history information, and the input $x_{t_j}$, which represents the current information. The hidden layer is decided by a $\mathbf{GRU}$ mechanism\cite{chung2014empirical}, where the reset gate controls how much of the past information to forget and the update gate determines how much new information to incorporate, using this gating mechanism to adaptively choose history information. Specifically, it computes the hidden state $l_j$ as:
\begin{equation}
\begin{aligned}
r_j &= \sigma(W_r x_{t_{j}} + U_r l_{j-1} + b_r), \\
z_j &= \sigma(W_z x_{t_{j}} + U_z l_{j-1} + b_z), \\
\tilde{l}_j &= \tanh(W_l x_{t_{j}} + U_l (r_j \odot l_{j-1}) + b_l), \\
l_j &= (1 - z_j) \odot l_{j-1} + z_j \odot \tilde{l}_j,
\end{aligned}
\end{equation}
where $r_j$, $z_j$, and $\tilde{l}_j$ are the reset gate, update gate, and candidate hidden state, respectively. $W_*, U_*$ are learnable weight matrices, $b_*$ are bias vectors, $\sigma$ denotes the sigmoid function, and $\odot$ represents element-wise multiplication.

% The Appendix \eqref{ablation_study} compiles ablation studies that compare the performance of the original MHP with the enhanced version incorporating both mechanisms, providing a detailed analysis of their individual contributions. There also compares several other Time-Scaling and Information Embedding methods like linear scaling and advanced information concatenate using dual gating, and finally $\mathbf{GRU}$ scaling and simple concat displayed as \eqref{time_scaling} and \eqref{eq:simple_concat} wins.

% In subsequent experiments, the MHP model architecture incorporated both mechanisms mentioned in this section without further reiteration. 

\begin{algorithm}[thb]
	\caption{Adaptive Mamba Hawkes Process (A-MHP)}
	\label{alg:enhanced_mhp}
	{\bf Input:}
	The temporal sequence $\{(t_1, k_1),(t_2,k_2),...,(t_L,k_L)\}$\\
	{\bf Output:} 
    The enhanced hidden state $\mathbf{h}$ and scaling factor $s_j$\\
	\begin{algorithmic}[1]
		\STATE ${x}_{t_j} \gets \text{Embed}(k_j) \in \mathbb{R}^{\text{d\_{model}}}$
		\STATE $\Delta \gets t_j - t_{j-1}$ // $\Delta$ records the temporal information
		\STATE // \textit{Time-Scaling Mechanism}
		\STATE $l_j \gets \text{GRU}({x}_{t_j}, l_{j-1})$
		\STATE $s_j \gets \text{Softplus}(\mathbf{W}_l^T l_j + b_l)$
		\STATE $\Delta' \gets \Delta \cdot s_j$
		\STATE // \textit{Dual-Channel State Transition}
		\STATE $\widetilde{\mathbf{B}}:(B,L,D) \gets \sigma([\mathbf{W}_c {x}_{t_j}; \mathbf{W}_\tau \Delta']) \odot \mathbf{B}_0$
		\STATE $\mathbf{C}:(B,L,D) \gets \text{Linear}_C({x}_{t_j})$ // $\mathbf{C}$ is time-dependent
		\STATE // \textit{SSM with Scaled Time}
		\STATE $\mathbf{A}:(L,D) \gets \text{Parameter}$ // learnable matrix
		\STATE $\overline{\mathbf{A}}, \overline{\mathbf{B}} \gets \text{Discretize}(\mathbf{A}, \widetilde{\mathbf{B}}, \Delta')$
		\STATE $\mathbf{h} \gets \text{SSM}(\overline{\mathbf{A}}, \overline{\mathbf{B}}, \mathbf{C}, \Delta')({x})$
		\RETURN $\mathbf{h}, s_j$
	\end{algorithmic}
\end{algorithm}

\subsection{Dual-Channel State Transition}\label{Dual-Channel}

Dual-Channel modifies $\mathbf{B}(t_j)$ to improve the selection mechanism by involving more information in data-dependent state transitions. This is realized by concatenating embedded events and time intervals:
% \begin{equation}
    \begin{align}
    \widetilde{\mathbf{B}}(t_j) &= \sigma([W_c x_{t_j}; W_\tau \Delta'_j]) \odot B_0, \label{eq:simple_concat}
    \end{align}
% \end{equation}
where $W_c \in \mathbb{R}^{\text{d\_{inner}} \times \text{d\_{state}}}$ is the content projection matrix, $W_\tau \in \mathbb{R}^{1 \times {\text{d\_{state}}}}$ is the temporal projection matrix, $B_0 \in \mathbb{R}^{\text{d\_{state}}}$ is the base transition vector implemented as learnable parameter in the model, $x_{t_j}$ is the embedded events, and $\Delta'_j$ is the scaled time interval. $\text{d\_{inner}}$ and $\text{d\_{state}}$ denote the internal feature dimension and state space dimension in the selective SSM respectively. The dual-channel design separates content processing (event-based via $W_c$) from temporal processing (time-based via $W_\tau$), ensuring that state transitions are influenced by both event semantics and temporal dynamics.

This design formalizes content-dependent temporal processing, a key objective of selective state spaces, while maintaining computational tractability. The sigmoid nonlinearity introduces negligible overhead versus linear layers.

\section{Experimental Verification and Analysis}\label{main_exp}
\subsection{Experiment Settings}

\subsubsection{\textbf{Datasets.}} To evaluate the performance of \modela~ and \modelb, we conduct experiments on five benchmark datasets: Synthetic Dataset\cite{zhang2020self}, Financial Transactions\cite{du2016recurrent}, StackOverFlow\cite{leskovec2016snap}, Retweet\cite{zhao2015seismic}, and MIMIC-II\cite{johnson2016mimic}. The statistic is shown in Table \ref{tab:statistic} in the appendix.

\subsubsection{\textbf{Baselines.}}
We compare the effectiveness of \modela~ and \modelb~ against the following baseline models:\\
\begin{itemize}
    \item \textbf{Recurrent Marked Temporal Point Process (RMTPP)} \cite{du2016recurrent} applies a Recurrent Neural Network (RNN) architecture to encode temporal information, creating a representation that supports sequential event prediction.
    \item \textbf{Neural Hawkes Process (NHP)}\cite{mei2017neural} constructs a neural self-modulating multivariate point process using a continuous-time LSTM, which allows the model to adaptively elevate or decrease event intensities based on historical patterns in asynchronous event sequences.
    \item \textbf{Self-attentive Hawkes Process (SAHP)}\cite{zhang2020self} leverages an attention mechanism to predict event sequences in a Hawkes process. It incorporates a hybrid positional encoding in its model construction, which fuses the temporal positional encoding and the absolute positional encoding.
    \item \textbf{Transformer Hawkes Process (THP)} \cite{zuo2020transformer} regards the time stamps as the position of the event vectors and applies absolute positional encoding in a Transformer architecture.
    \item \textbf{Attentive Temporal Point Process (AttnTPP)} \cite{bae2023meta} formulates TPP as Neural Processes (NPs) and learn local patterns with a cross-attention mechanism.
\end{itemize}
\subsubsection{\textbf{Metrics}}
We assessed model performance using three key metrics: log-likelihood (\textsl{LL}), accuracy (\textsl{Acc}), and Root Mean Square Error (\textsl{RMSE}). Log-likelihood indicates how well the model fits the observed data, accuracy captures the correctness of event type predictions, and Root Mean Square Error measures the precision of event time predictions.

\subsubsection{\textbf{Implementation}}

\iffalse
For \modela, we set the embedding dimension as described in Table~\ref{table:setting}. Additionally, all additional architectural hyperparameters are detailed in Table~\ref{table:mhp}. in appendix. For \modelb, the Mamba component adopts the same design as in \modela. In addition, in \modela's dual-channel design, the state dimension $\text{d\_state}$ is partitioned between the event and temporal channels to maintain dimensional consistency with SSMs. As event embeddings capture richer semantic information than scalar time intervals, we allocate two-thirds of $\text{d\_state}$ to the event channel and the remaining one-third to the temporal channel. The Transformer blocks follow the architecture of THP \cite{zuo2020transformer}. Likewise, we adhere to THP’s configuration, adjusting the loss function weights according to the specific characteristics of each dataset.
% To avoid NaN values during our training, we apply Softplus function and Clamp function to the temporal difference $\Delta$, along with gradient clipping, label smoothing regularization, and enhanced numerical stability protections for robust training.
All experiments are performed on a GPU RTX 3090 with 24GB memory.
\fi

For \modela, the embedding dimension is configured as specified in Table~\ref{table:setting} in the appendix, while all additional architectural hyperparameters are provided in Table~\ref{table:mhp} in the appendix. For \modelb, the Mamba component adopts the same design as in \modela. In addition, in \modela's dual-channel design, the state dimension $\text{d\_state}$ is partitioned between the event and temporal channels to maintain dimensional consistency with SSMs. As event embeddings capture richer semantic information than scalar time intervals, we allocate two-thirds of $\text{d\_state}$ to the event channel and the remaining one-third to the temporal channel. The Transformer blocks follow the architecture of THP \cite{zuo2020transformer}. Likewise, we adhere to THP’s configuration, adjusting the loss function weights according to the specific characteristics of each dataset.
% To avoid NaN values during our training, we apply Softplus function and Clamp function to the temporal difference $\Delta$, along with gradient clipping, label smoothing regularization, and enhanced numerical stability protections for robust training.
All experiments are performed on a GPU RTX 3090 with 24GB memory.

\iffalse
\begin{table}[thb]
	%\begin{minipage}{0.48\textwidth}
	\centering
	\caption{Hyper-parameter settings of different datasets.}
	\label{table:setting}
	\footnotesize
	\begin{tabular}{l|c|c|c|c|c}
		\toprule
		\textbf{Parameter} & d\_model & Learning Rate & Batch Size & $\beta$ & $\gamma$ \\
		\midrule
		\textbf{Financial} & 128 & 1e-4 & 1 & 1.0 & 1e-4 \\ 
		\textbf{SO} & 512 & 1e-4 & 4 & 2.0 & 8e-5 \\ 
		\textbf{Synthetic} & 64 & 1e-4 & 4 & 1.0 & 1e-4 \\ 
		\textbf{Retweet} & 64 & 1e-3 & 16 & 0.2 & 4e-4 \\ 
		\textbf{Mimic-II} & 64 & 5e-4 & 1 & 1.6 & 8e-5 \\ 
		\bottomrule
	\end{tabular}
\end{table}
\fi

\iffalse
\begin{table}[thb]
	%\begin{minipage}{0.48\textwidth}
	\centering
	\caption{Architecture of proposed \modela.}
	\label{table:mhp}
	\footnotesize
	\begin{tabular}{@{}ccccccc@{}}%{lp{4cm}}
		\toprule
		\text{d\_inner} & \text{d\_state} & \text{d\_conv} & \text{expand factor}&\text{n\_layers} & \text{dt\_rank} & \text{B\_split}\\
		\midrule
		\text{2$\times$d\_model} & 16 & 4 & 2 & 4 & $\text{d\_model}/16$ & \text{2 : 1}\\ 
		\bottomrule
	\end{tabular}
	
\end{table}
\fi

\subsection{Experimental Results}

\begin{figure}[thb]
    \centering
    \begin{subfigure}{0.15\textwidth}
        \centering
        \includegraphics[width=\linewidth]{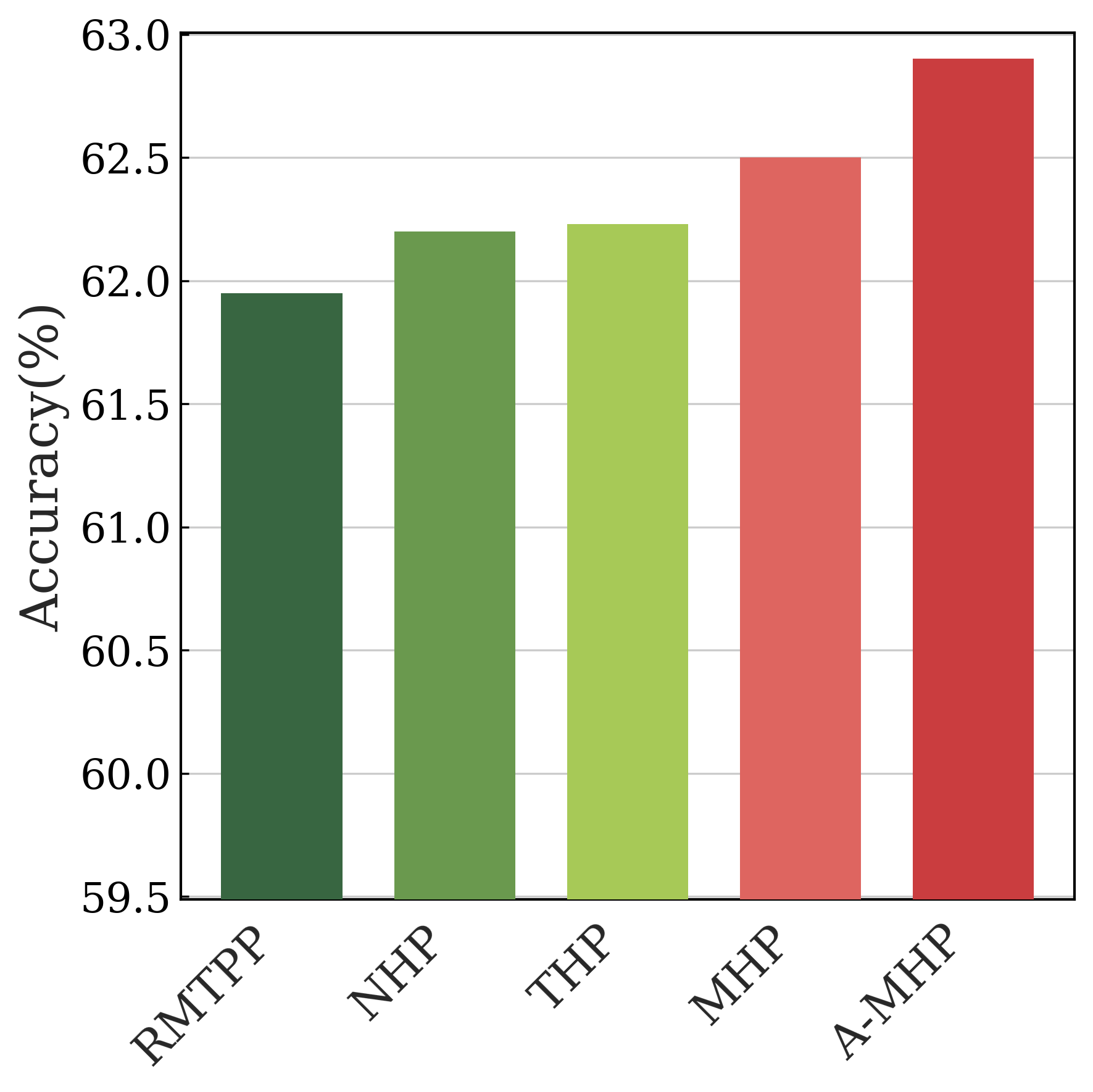}
        \caption{Financial}
        \label{fig:acc_fin}
    \end{subfigure}
    \hspace{0.005\textwidth}
    \begin{subfigure}{0.15\textwidth}
        \centering
        \includegraphics[width=\linewidth]{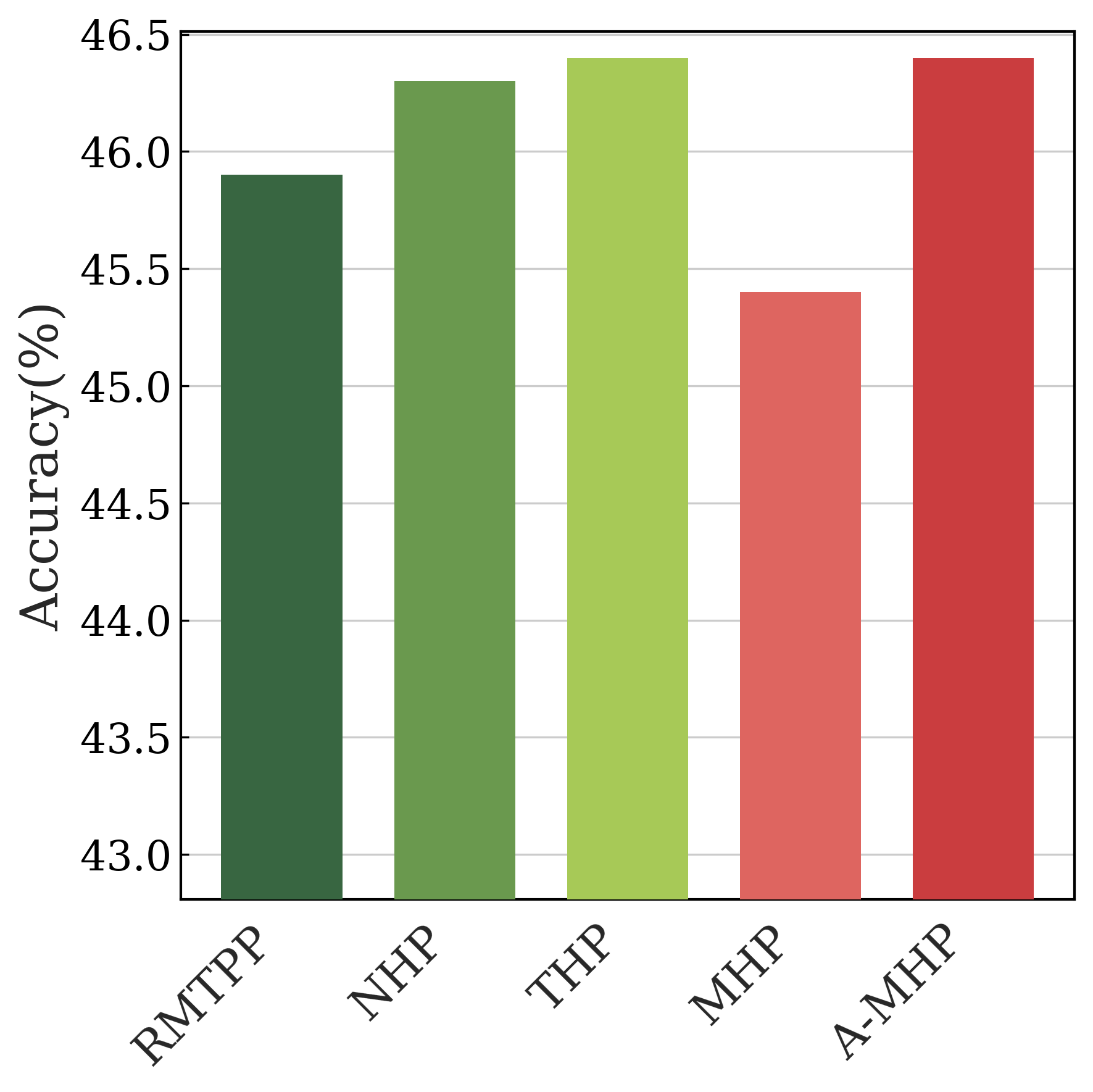}
        \caption{SO}
        \label{fig:acc_so}
    \end{subfigure}
    \hspace{0.005\textwidth}
    \begin{subfigure}{0.15\textwidth}
        \centering
        \includegraphics[width=\linewidth]{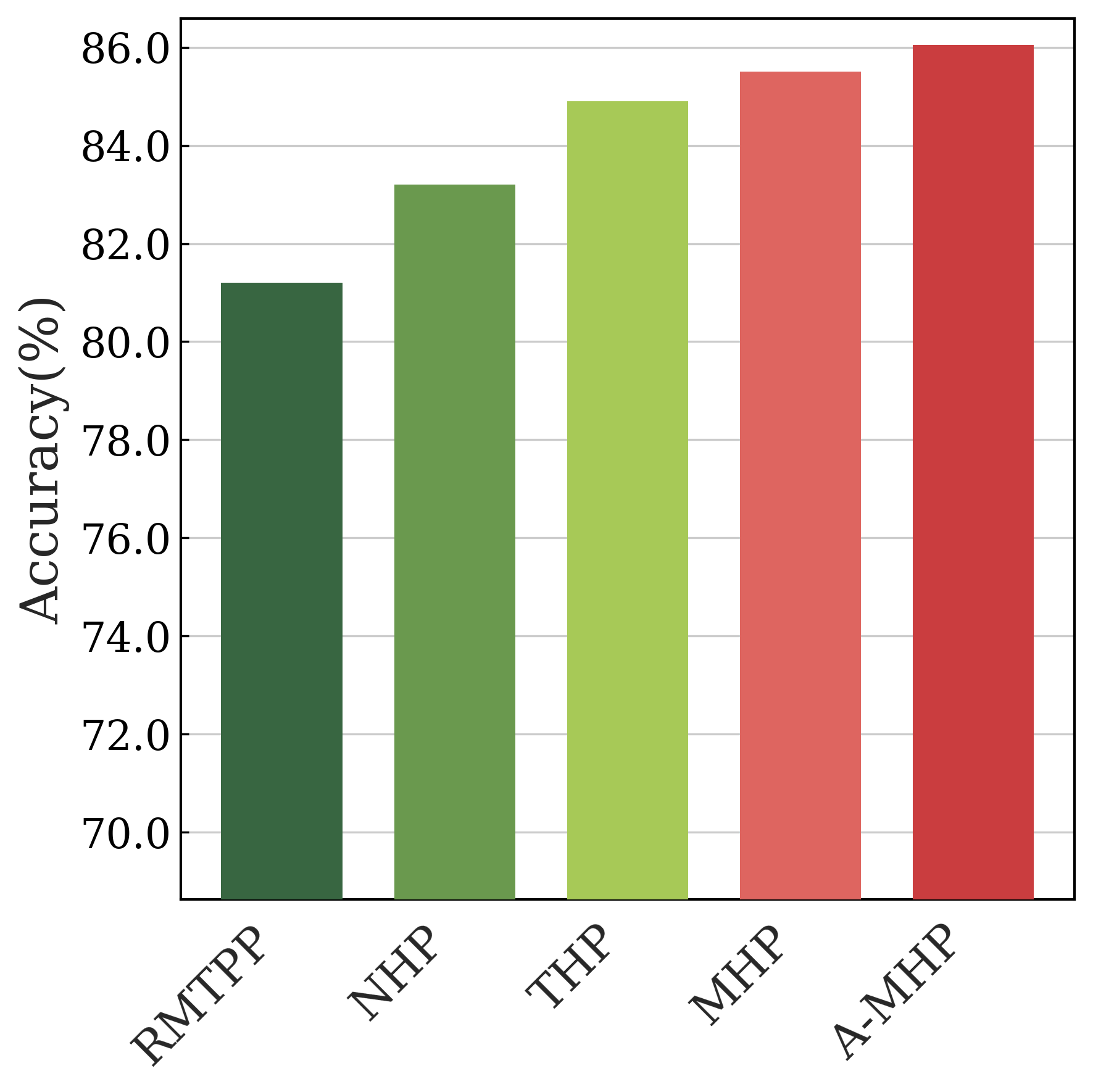}
        \caption{MIMIC-II}
        \label{fig:acc_mimic}
    \end{subfigure}
    \caption{Accuracy comparison on three datasets.}
    \label{fig:acc}
\end{figure}

\begin{figure}[thb]
    \centering
    \begin{subfigure}{0.15\textwidth}
        \centering
        \includegraphics[width=\linewidth]{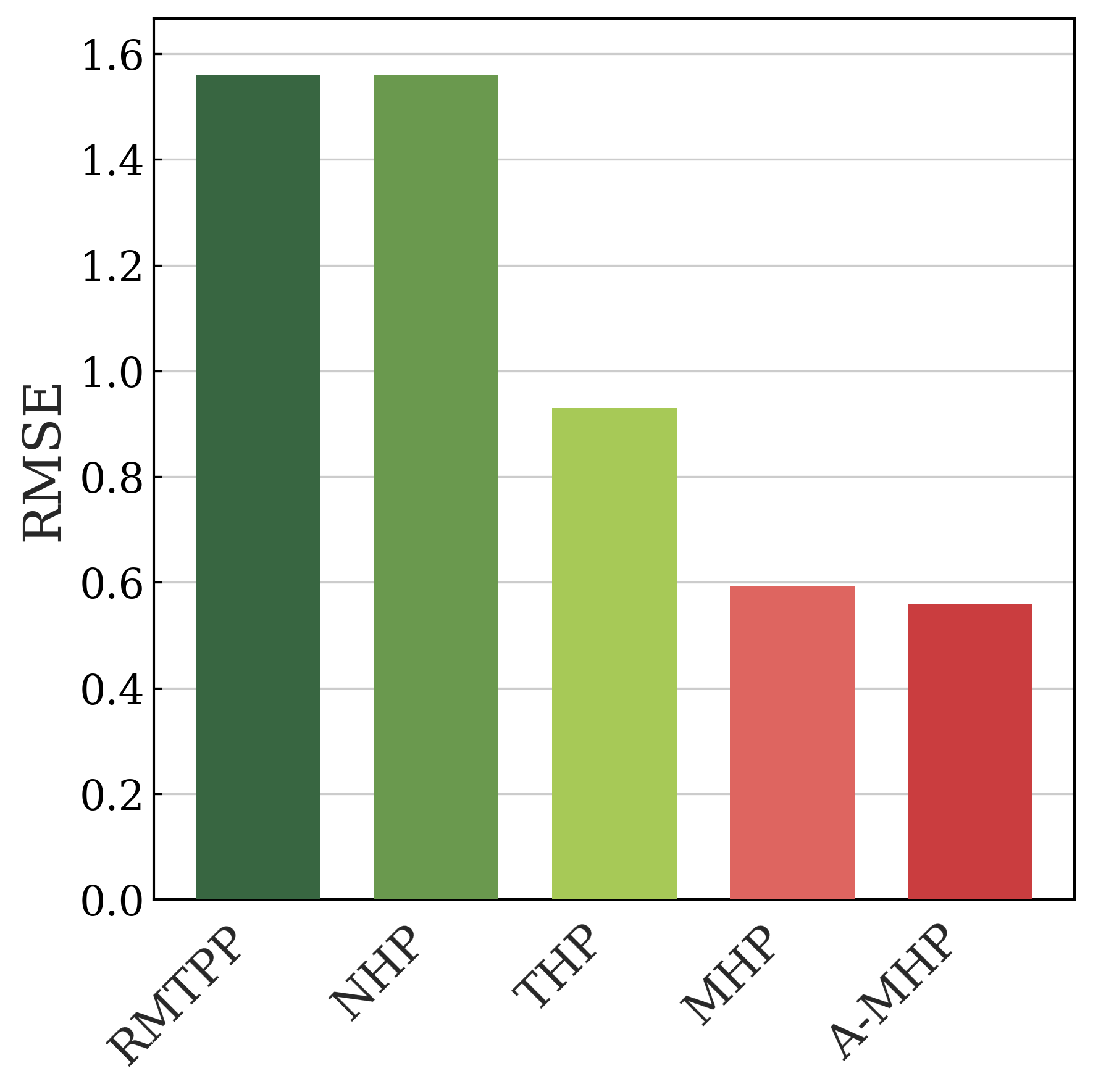}
        \caption{Financial}
        \label{fig:rmse_fin}
    \end{subfigure}
    \hspace{0.005\textwidth}
    \begin{subfigure}{0.15\textwidth}
        \centering
        \includegraphics[width=\linewidth]{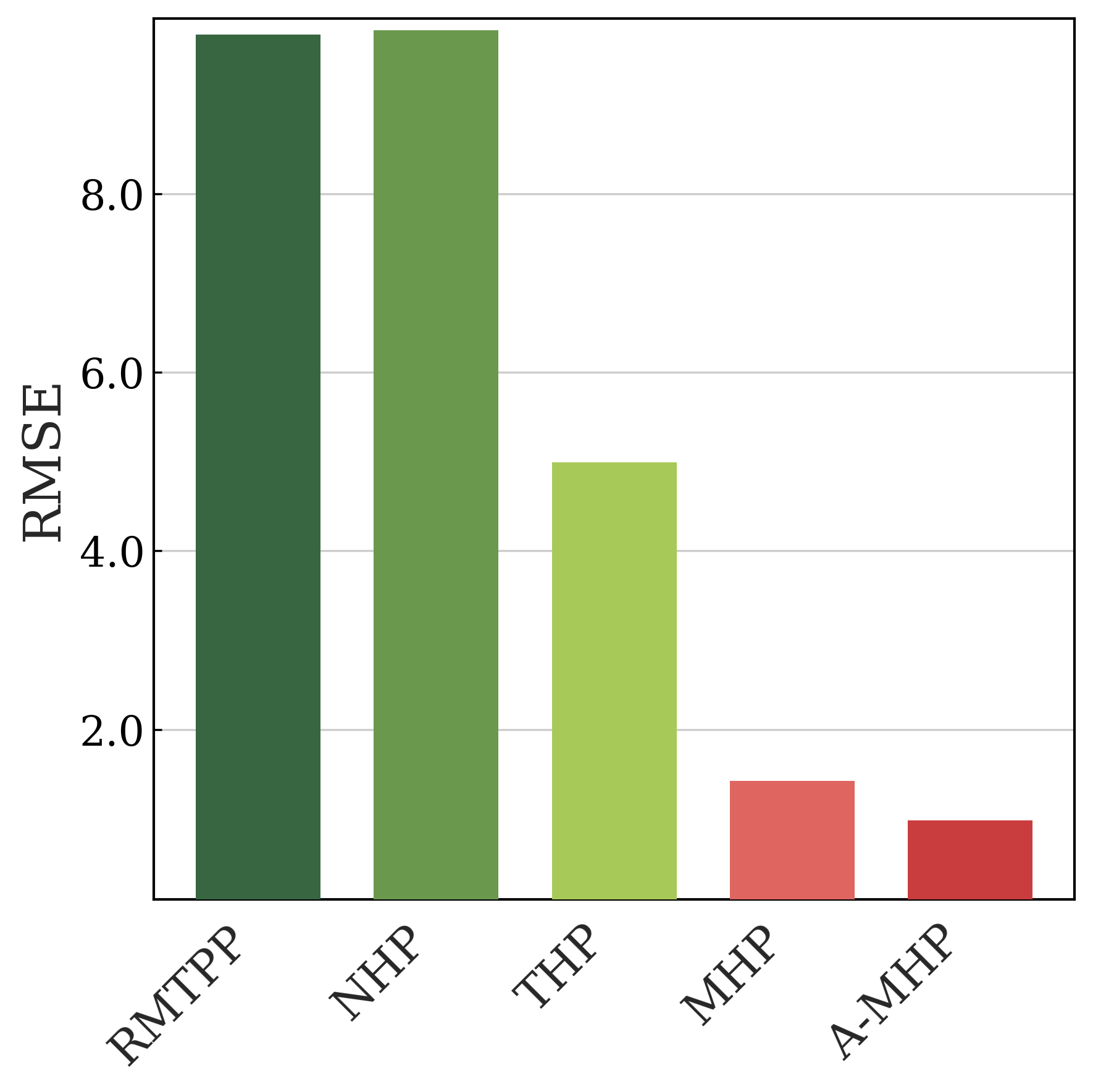}
        \caption{SO}
        \label{fig:rmse_so}
    \end{subfigure}
    \hspace{0.005\textwidth}
    \begin{subfigure}{0.15\textwidth}
        \centering
        \includegraphics[width=\linewidth]{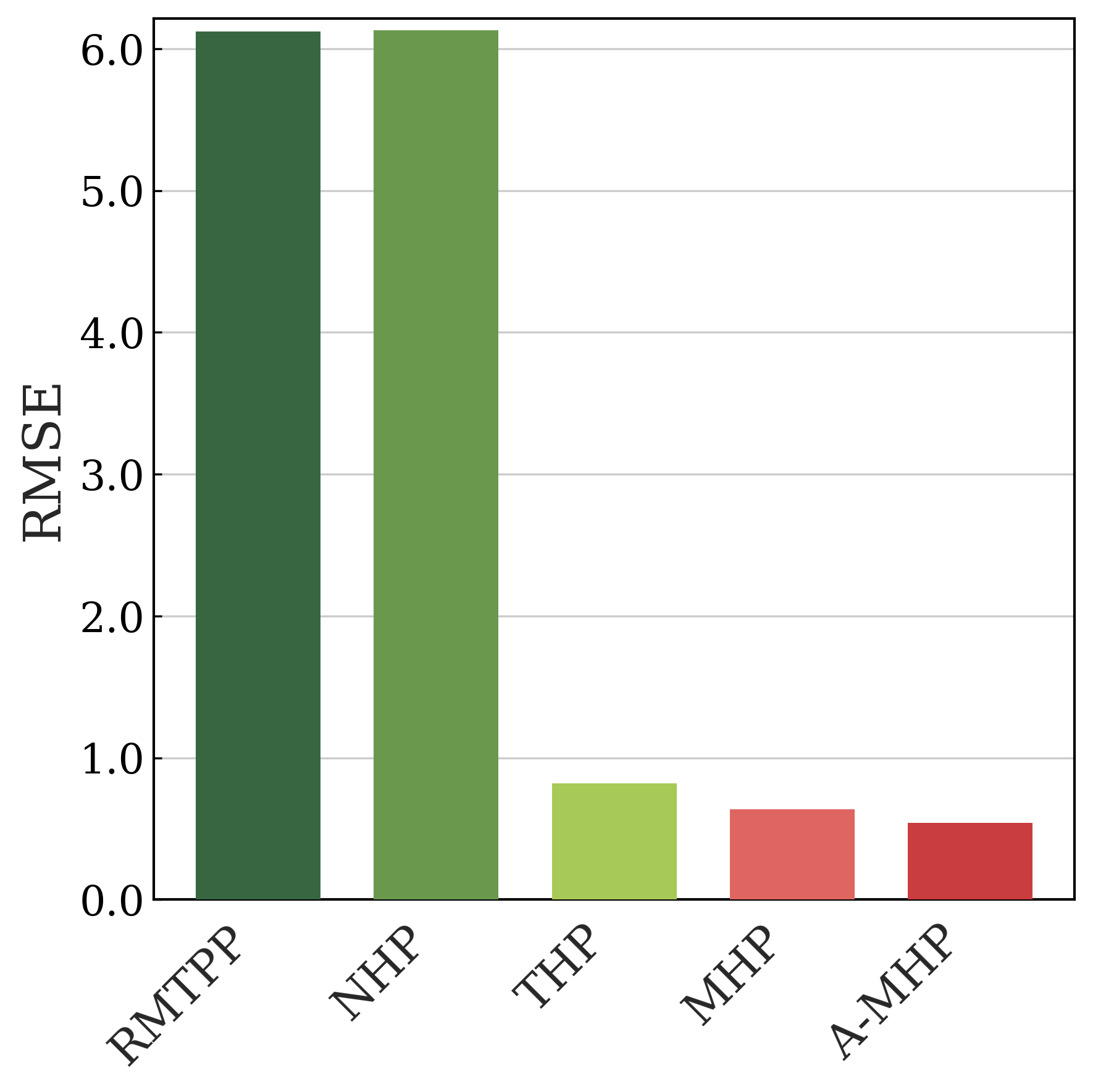}
        \caption{MIMIC-II}
        \label{fig:rmse_mimic}
    \end{subfigure}
    \caption{RMSE comparison on three datasets.}
    \label{fig:rmse}
\end{figure}

This section presents a comprehensive evaluation of the proposed \modela~ and \modelb~ models against the baselines. 
We report the Log-Likelihood (Table \ref{tab:ll}), event type prediction Accuracy (Figure \ref{fig:acc}), and Root Mean Square Error (Figure \ref{fig:rmse}) for event time prediction across five datasets. 
% The Log-Likelihood (\textsl{LL}) scores are compared in Table \ref{tab:ll}, the next event type prediction Accuracy (\textsl{Acc}) is detailed in Figure \ref{fig:acc}, and the Root Mean Square Error (\textsl{RMSE}) for event time prediction is shown in Figure \ref{fig:rmse}. 
% A high-level overview reveals that \modelb~ achieves the top performance across all five datasets, establishing a new state-of-the-art. \modela~ also shows highly competitive results, often surpassing all previous baselines. 
% Beyond these aggregate observations, the following subsections will dive into a nuanced analysis of these results, examining the architectural and mechanistic advantages of our models.
In what follows, we dive into a nuanced analysis of these results, examining the architectural and mechanistic advantages of our models.

\textbf{Architectural Evolution Directly Drives Performance Gains.}
The progression of model performance from RMTPP, through SAHP and THP, to our proposed \modela/\modelb~ reflects a clear architectural evolution in temporal point process modeling.
RMTPP, grounded in a simple recurrent neural network, is constrained by recurrent bottlenecks, which hinder its ability to capture long-range and complex event dependencies.
SAHP, introducing self-attention, allows the model to directly account for all historical events and yields notable improvements on metrics such as \textsl{LL} and \textsl{RMSE}.
Continued with THP, which adopts the full Transformer architecture. Its superior performance, particularly in \textsl{LL}, highlights the effectiveness of multi-head self-attention in modeling global dependencies across entire event sequences.
Collectively, these developments emphasize the central role of attention mechanisms in temporal point processes, while also revealing the scalability limitations of softmax-based attention that our model is designed to overcome.

\begin{table}[thb]
\centering
\caption{Log-Likelihood result of all baseline methods across five datasets.}
\label{tab:ll}
\begin{adjustbox}{max width=0.5\textwidth}
\setlength{\tabcolsep}{3pt} % 调整列间距
\begin{tabular}{l||ccccc}
\toprule
Method & Financial & SO & Synthetic & Retweet & MIMIC-II \\
\midrule
RMTPP     & -3.89 & -2.60 & -1.33 & -5.99 & -1.35 \\
NHP      & -3.60 &  -2.55 & - & -5.60 & -1.38 \\
SAHP      & - & -1.86 & 0.59 & -4.56 & -0.52 \\
THP      & -1.11 & -0.04 & 0.79 & -2.04 & 0.48 \\
AttnTPP      & -4.26 & -2.97 & -1.20 & -5.99 & -1.05 \\
\midrule
\rowcolor[gray]{0.9}\modela & \underline{1.25} & \underline{0.40} & \underline{1.05} & \underline{0.83} & \underline{1.46} \\
\rowcolor[gray]{0.9}\modelb & \textbf{1.33} & \textbf{0.42} & \textbf{1.08} & \textbf{1.07} & \textbf{1.62} \\
\bottomrule
\end{tabular}
\end{adjustbox}
\end{table}

\textbf{Mamba Architecture Excels in Long-Range and Variable-Length Dependency Modeling.}
\modela~ and \modelb~ demonstrate superior performance on datasets characterized by long and highly variable sequence lengths.
% The Financial dataset and Retweet dataset represent significant challenges for traditional models. 
% Transformer-based models like THP, despite their power, can struggle with information dilution in such extremely long sequences. 
While attention mechanisms can struggle with context dilution in highly variable-length sequences, the SSM's hidden state provides a continuous, fixed-size representation of the entire event history. This design makes the model's context processing inherently invariant to sequence length and resolution, allowing it to generalize seamlessly from short to extremely long sequences without performance degradation. 
% In contrast, the Mamba backbone in our models, with its linear-time complexity and selective state space mechanism, efficiently captures long-range dependencies. 
This is evidenced by the substantial gains in \textsl{LL} and \textsl{Acc} on the Financial dataset and Retweet dataset. Furthermore, the low standard deviation of the Synthetic and MIMIC-II datasets explains why models perform reasonably well there, as the temporal patterns are more consistent.
% Furthermore, the low standard deviation of Synthetic and mimic explains why all models perform reasonably well there, as the temporal patterns are more consistent and shorter-range. Our model's advantage lies in its ability to handle not just short, structured sequences but also the long, chaotic ones where others falter.

\textbf{Performance Margins Depend on Dataset Complexity.}
On datasets with fewer event types and shorter sequences, such as Synthetic, the relative improvements of our models over strong baselines are less pronounced. 
In these simpler settings, most modern architectures can capture the limited temporal variability with reasonable accuracy, which narrows the performance margins. 
By contrast, in more complex datasets like Financial, which involve long-range dependencies and heterogeneous event dynamics, the advantages of \modela~ and particularly \modelb~ become much more evident.

\textbf{Overall Superiority of \modela~ and \modelb.}
Taken together, these results consistently show that \modelb~ achieves the best overall performance across all datasets and metrics, with \modela~ also performing competitively. 
This confirms that the proposed architectural not only overcome the limitations of existing approaches but also generalize robustly across diverse temporal domains.

\begin{table*}[htb]
\centering
\caption{Ablation study results on the Dual-Channel and Time-Scaling components.}
\label{tab:ablation}
\begin{adjustbox}{max width=\textwidth}
\renewcommand{\arraystretch}{1.3} % 调整行高
\begin{tabular}{l||ccccccccccccccc}
\toprule
\multirow{2}{*}{Method} & \multicolumn{3}{c}{Financial} & \multicolumn{3}{c}{SO} & \multicolumn{3}{c}{Synthetic} & \multicolumn{3}{c}{Retweet} & \multicolumn{3}{c}{MIMIC-II} \\
\cmidrule(lr){2-4} \cmidrule(lr){5-7} \cmidrule(lr){8-10} \cmidrule(lr){11-13} \cmidrule(lr){14-16}
 & \textsl{LL} & \textsl{Acc} & \textsl{RMSE} & \textsl{LL} & \textsl{Acc} & \textsl{RMSE} & \textsl{LL} & \textsl{Acc} & \textsl{RMSE} & \textsl{LL} & \textsl{Acc} & \textsl{RMSE} & \textsl{LL} & \textsl{Acc}  & \textsl{RMSE} \\
\midrule
\modela & 1.2469 & 0.6245 & 0.5956 & 0.3967 & 0.4539 & 1.4275 & 1.0516 & 0.3813 & \textbf{0.1817} &  0.8285 & 0.5605 & \textbf{5.4275} &  1.4607 & 0.8550 & 0.6353 \\
\textcolor{YellowOrange}{\textbf{DC}}  & 1.3153 & 0.6276 & 0.5868 & 0.4131 &  0.4639 & 1.3335 & 1.0748 & 0.3813 & 0.1861 & 1.0514 & 0.6086 & 5.5523 & 1.5959 & 0.8514 & 0.6083 \\
\textcolor{LimeGreen}{\textbf{TS}} &  1.3250 & 0.6283 & 0.5883 & 0.4154 & 0.46284 & 1.1061 & 1.0682 & 0.3813 & 0.2016 & 1.0426 & 0.6080 & 5.5523 & 1.6056 & 0.8572 & 0.5511 \\
\midrule
% \rowcolor[gray]{0.9}MHP & 0.00 & 0.00 & 0.00 & 0.00 & 0.00 & 0.00 & 0.00 & 0.00 & 0.00 & 0.00 & 0.00 & 0.00 & 0.00 & 0.00 & 0.00 \\
\rowcolor[gray]{0.9}\textcolor{YellowOrange}{\textbf{DC}} + \textcolor{LimeGreen}{\textbf{TS}} & \textbf{1.3272} & \textbf{0.6290} & \textbf{0.5631} & \textbf{0.4233} & \textbf{0.4646} & \textbf{0.9835} & \textbf{1.0825} & \textbf{0.3813} & 0.1995 & \textbf{1.0708} & \textbf{0.6086} & 5.5523 & \textbf{1.6181} & \textbf{0.8605} & \textbf{0.5400} \\
\bottomrule
\end{tabular}
\end{adjustbox}
\end{table*}

\subsection{\modela~ V.S. THP}
To further highlight the strengths of the proposed model, we conduct a direct comparison with the transformer-based THP model.
We focus on two key dimensions: (i) \textbf{Efficiency}, which measures computational throughput across datasets, and (ii) \textbf{Effectiveness}, which evaluates predictive performance and model scalability.

\subsubsection{\textbf{Efficiency}}
\iffalse
\begin{table}[thb]
\centering
\caption{Throughput (events/s) comparison and relative speedup between THP and \modela.}
\label{tab:throughput}
\begin{tabular}{l|cc|c}
\toprule
Dataset & THP & \modela & Speedup \\
\midrule
Financial & 1167.40 & 5088.59 & 4.36$\times$ \\
SO &  20203.13 & 21608.38 & 1.07$\times$ \\
Synthetic & 8064.19 & 15562.60 & 1.93$\times$ \\
Retweet &  12733.84 & 78983.22 & 6.20$\times$ \\
MIMIC-II & 302.12 & 494.26 & 1.64$\times$ \\
\bottomrule
\end{tabular}
\vspace{-0.5em}
\end{table}
\fi

To assess the efficiency of proposed \modela~ against the Transformer-based THP, we report throughput (events per second) and relative speedup across five benchmark datasets (Table~\ref{tab:throughput} in the appendix). Overall, \modela~ achieves consistently higher training throughput, with the most significant gains observed on the Financial and Retweet datasets, where the speedup reaches $4.36\times$ and $6.20\times$, respectively. This improvement stems from the architectural advantages of the Mamba backbone: unlike Transformers, which rely on quadratic self-attention operations, Mamba employs selective state-space dynamics with linear-time complexity. This design substantially reduces computational overhead, particularly in handling long or irregular event sequences.

\subsubsection{\textbf{Effectiveness}}

\begin{figure}[thb]
    \centering
    \begin{subfigure}{0.15\textwidth}
        \centering
        \includegraphics[width=\linewidth]{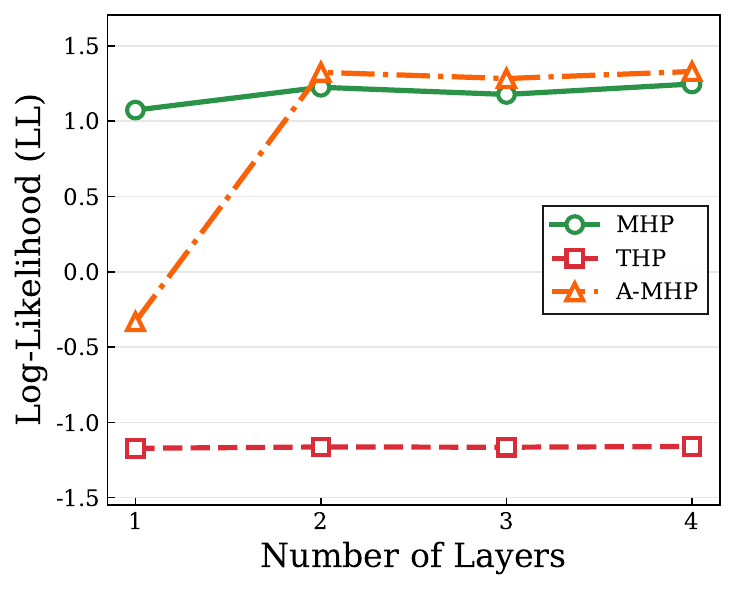}
        \caption{Financial}
        \label{fig:layer_fin}
    \end{subfigure}
    \hspace{0.005\textwidth}
    \begin{subfigure}{0.15\textwidth}
        \centering
        \includegraphics[width=\linewidth]{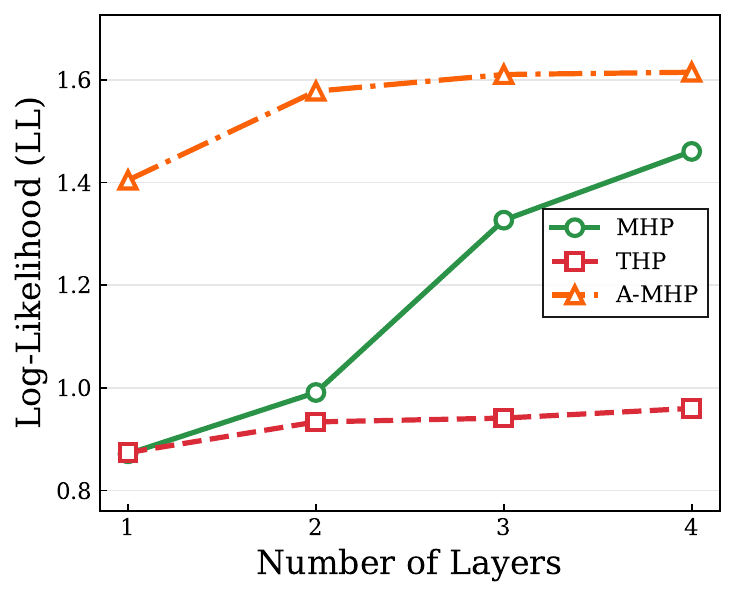}
        \caption{MIMIC-II}
        \label{fig:layer_mimic}
    \end{subfigure}
    \hspace{0.005\textwidth}
    \begin{subfigure}{0.15\textwidth}
        \centering
        \includegraphics[width=\linewidth]{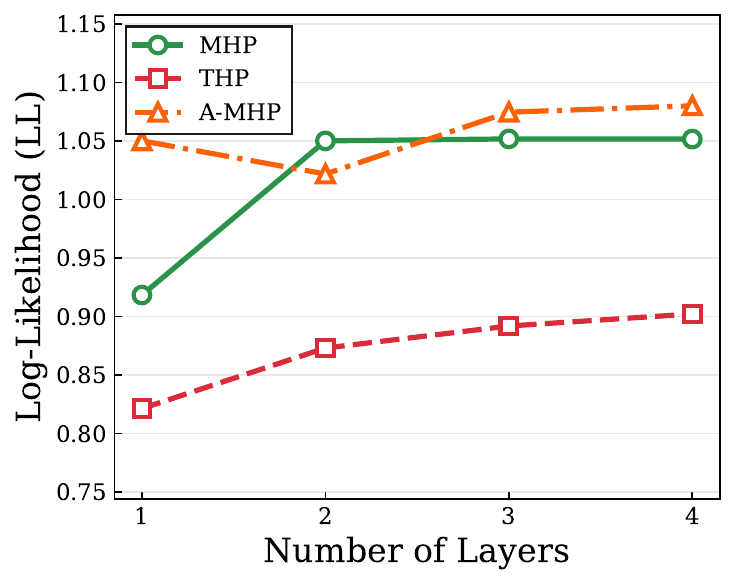}
        \caption{Synthetic}
        \label{fig:layer_synthetic}
    \end{subfigure}
    \caption{Performance of \modela, THP, and \modelb~ across different layer counts, evaluated by Log-Likelihood.}
    \label{fig:layer}
\end{figure}

To elucidate the inherent architectural advantages of proposed models, Figure \ref{fig:layer} compares the scaling behaviors of \modela~ and \modelb~ against THP, using network depth (1–4 layers) as a controlled variable to assess performance progression. As revealed by the results, all models exhibit performance gains with increasing layers, confirming the fundamental importance of network depth for capturing complex temporal dependencies. While THP improves slightly with increasing depth, its rate of growth quickly plateaus, indicating that its ability to leverage additional layers is limited due to its pure reliance on the Transformer architecture. In contrast, \modela~ and \modelb~ exhibit stronger depth scalability, enabling them to more effectively model complex dependencies as the network becomes deeper.

\subsection{Ablation Study}

To validate the contribution of each proposed mechanism, we design two ablation variants that isolate individual components of \modelb~ (Table \ref{tab:ablation}):

\textbf{MHP+Time-Scaling (\textcolor{LimeGreen}{TS})}: Retains the complete time-scaling mechanism  in \eqref{time_scaling} but uses simple linear projection for state transition. The scaled time intervals $\Delta'_j$ are used throughout SSM computations, and time predictions are divided by scaling factors.

\textbf{MHP+Dual-Channel (\textcolor{YellowOrange}{DC})}: Implements the dual-channel state transition in \eqref{eq:simple_concat} but removes time scaling entirely. Original time intervals $\Delta_j$ are used in SSM, and time predictions are computed directly without scaling adjustments.

% Both variants maintain identical network architectures, hyperparameters, and training procedures as \modelb~ to ensure fair comparison. The base \modela~ model serves as the control baseline.
% Based on the ablation results presented in Table \ref{tab:ablation}, we draw the following conclusions regarding the contribution of each component:

\subsubsection{\textbf{ Individual Effectiveness of \textcolor{YellowOrange}{DC} and \textcolor{LimeGreen}{TS}}} Both the \textcolor{YellowOrange}{\textbf{DC}} and \textcolor{LimeGreen}{\textbf{TS}} mechanisms individually contribute to performance improvements over the \modela~ model across most datasets and evaluation metrics. The \textcolor{YellowOrange}{\textbf{DC}} variant shows consistent gains in \textsl{LL} and \textsl{Acc}, particularly evident in the Retweet dataset, where the \textsl{Acc} improves from 56.05\% to 60.86\%. The \textcolor{LimeGreen}{\textbf{TS}} variant, on the other hand, delivers more pronounced improvements in time prediction accuracy, as seen by the lower \textsl{RMSE} on SO and MIMIC-II. This confirms its role in adapting to multi-timescale dynamics through resolution-aware interval scaling.

\subsubsection{\textbf{Complementary Nature and Synergistic Effects}} The full \modelb~ model achieves the best overall performance, frequently surpassing either variant used alone. This indicates that the two components are largely complementary. For instance, on the Financial dataset, while both \textcolor{YellowOrange}{\textbf{DC}} and \textcolor{LimeGreen}{\textbf{TS}} individually achieve an \textsl{LL} of 1.3153 and 1.3250, respectively, their combination pushes the performance further to 1.3272 while also yielding the best \textsl{RMSE}. This synergy suggests that the benefits of refined state transition (\textcolor{YellowOrange}{\textbf{DC}}) and adaptive time representation (\textcolor{LimeGreen}{\textbf{TS}}) can be effectively combined.

On the Synthetic and Retweet datasets, \modela~ alone performs best while additional modules do not yield further gains, and the metrics (\textsl{ACC}/\textsl{RMSE}) show repeated values due to rounding.
This can be attributed to the simplicity of these datasets, as they involve fewer event types and relatively short sequences, reducing the benefit of additional modeling complexity.

% Table \ref{tab:ablation} demonstrates that not only does each individual mechanism effectively contribute to the improvement of the three key metrics, but their combined effect is even more pronounced. The results show obvious improvements in the performance of the model. 
% A minor issue arose solely with the calculation of the retweet log-likelihood. We attribute this to potential gradient instability, likely caused by the structure of the data itself. The log-likelihood approached zero after clamping extreme values. Nonetheless, we consider this to be an engineering issue in the implementation, which does not affect the conclusions regarding the model's performance.

% The component interaction analysis in Figure~\ref{fig:component_interaction} reveals synergistic effects between the two mechanisms, where the combined implementation achieves performance gains beyond the sum of individual contributions.
\section{Conclusion}\label{conclusion}
In this work, we proposed the Mamba Hawkes Process (\modela), a novel model that integrates the selective state space framework of Mamba with temporal point processes to efficiently capture long-range dependencies in asynchronous event sequences with linear complexity. Theoretically, \modela~ generalizes classical Hawkes processes and certain neural point processes, providing a unified modeling perspective. We further introduced \modelb, which enhances adaptability through a Time-Scaling Mechanism and Dual-Channel State Transition, enabling more nuanced handling of heterogeneous event dynamics. Extensive experiments across synthetic and real-world datasets demonstrate that \modela~ and \modelb~ consistently outperform state-of-the-art baselines in both event type and time prediction, with particularly strong performance on long sequences. This work establishes a foundational integration of state space models into TPPs and opens avenues for future extensions, such as spatiotemporal modeling, causal interpretability, and resource-efficient deployment.
\begin{acks}
This work was partially supported by Shenzhen Science and Technology Program under Grant RCBS20221008093331068, NSFC under Grant 62302421, Basic and Applied Basic Research Fund in Guangdong Province under Grant 2025A1515010439,  and the Guangdong Provincial Key Laboratory of Big Data Computing, The Chinese University of Hong Kong, Shenzhen.
\end{acks}

\bibliographystyle{ACM-Reference-Format}
\bibliography{reference}

% \appendix
\newpage
\appendix

\section{Dataset details}

\begin{table}[tbh]
    \centering
     \caption{Dataset Statistics}
    \begin{tabular}{l|ccccc}
    \toprule
    Info &  Financial  & SO & Synthetic & Retweet & MIMIC-II  \\
    \midrule
    Len (Avg.)     & 2074.0 & 72.4 & 60.3 & 108.8 & 3.7\\
    Len (Std.)     & 1245.0 & 43.7 & 23.1 & 54.4 & 2.3\\
    \#Type     & 2 & 22 & 5 & 3 & 75\\
    \bottomrule
    \end{tabular}
    %\caption{Dataset Statistics}
    \label{tab:statistic}
\end{table}

\section{Hyper-parameter and architecture details}\label{sec:tables}
\begin{table}[thb]
	%\begin{minipage}{0.48\textwidth}
	\centering
	\caption{Hyper-parameter settings of different datasets.}
	\label{table:setting}
	\footnotesize
	\begin{tabular}{l|c|c|c|c|c}
		\toprule
		\textbf{Parameter} & d\_model & Learning Rate & Batch Size & $\beta$ & $\gamma$ \\
		\midrule
		\textbf{Financial} & 128 & 1e-4 & 1 & 1.0 & 1e-4 \\ 
		\textbf{SO} & 512 & 1e-4 & 4 & 2.0 & 8e-5 \\ 
		\textbf{Synthetic} & 64 & 1e-4 & 4 & 1.0 & 1e-4 \\ 
		\textbf{Retweet} & 64 & 1e-3 & 16 & 0.2 & 4e-4 \\ 
		\textbf{Mimic-II} & 64 & 5e-4 & 1 & 1.6 & 8e-5 \\ 
		\bottomrule
	\end{tabular}
\end{table}

\begin{table}[thb]
	%\begin{minipage}{0.48\textwidth}
	\centering
	\caption{Architecture details of proposed \modela.}
	\label{table:mhp}
	\footnotesize
	\begin{tabular}{@{}ccccccc@{}}%{lp{4cm}}
		\toprule
		\text{d\_inner} & \text{d\_state} & \text{d\_conv} & \text{expand factor}&\text{n\_layers} & \text{dt\_rank} & \text{B\_split}\\
		\midrule
		\text{2$\times$d\_model} & 16 & 4 & 2 & 4 & $\text{d\_model}/16$ & \text{2 : 1}\\ 
		\bottomrule
	\end{tabular}
	
\end{table}

\section{Efficiency Comparisons}

\begin{table}[thb]
\centering
\caption{Throughput (events/s) comparison and relative speedup between THP and \modela.}
\label{tab:throughput}
\begin{tabular}{l|cc|c}
\toprule
Dataset & THP & \modela & Speedup \\
\midrule
Financial & 1167.40 & 5088.59 & 4.36$\times$ \\
SO &  20203.13 & 21608.38 & 1.07$\times$ \\
Synthetic & 8064.19 & 15562.60 & 1.93$\times$ \\
Retweet &  12733.84 & 78983.22 & 6.20$\times$ \\
MIMIC-II & 302.12 & 494.26 & 1.64$\times$ \\
\bottomrule
\end{tabular}
\vspace{-0.5em}
\end{table}

\section{Proof}\label{sec:proof}
	\subsection{\textbf{Proposition 1}}

	\begin{proof}
		The setting shows a linear system driven by a sum of Dirac impulses. By the Green's function $G(t)=e^{At}\mathbf{1}_{t\ge 0}$, we get
		\begin{equation}
			h(t)=\int_0^t e^{A(t-\tau)} B\, u(\tau)\, d\tau = \sum_{j: t_j<t} e^{A(t-t_j)} B \mathbf{k}_j.
		\end{equation}
		With diagonal $A$, the $r$-th coordinate is 
		\begin{equation}
		 		h_r(t) = \sum_{j: t_j < t} B_{r,k_j} e^{-\omega_r (t-t_j)}.   
		\end{equation}

\noindent Substituting into intensity function and grouping by type produces the stated multi-exponential kernels.
\begin{align}
		\lambda_k(t) &= \mu_k + \sum_{r=1}^R c_{kr} \sum_{j: t_j < t} B_{r,k_j} e^{-\omega_r (t-t_j)}
		\nonumber \\&= \mu_k + \sum_{k'=1}^K \sum_{j: k_j=k'} \left( \sum_{r=1}^R c_{kr} B_{r k'} e^{-\omega_r (t-t_j)} \right),
\end{align}
which is a Hawkes process with multi-exponential kernels $\psi_{kk'}(u)=\sum_{r} c_{kr} B_{r k'} e^{-\omega_r u}$.
\end{proof}
	
\begin{remark}
If one uses ZOH with piecewise constant $x(t)\equiv \tilde{x}_i$ on $[t_i,t_{i+1})$, the discrete update carries a factor $A^{-1}(e^{A\Delta_i}-I)$, which depends on $\Delta_i$. Unrolling through time, $h_r(t)$ becomes a weighted sum of exponentials with interval-dependent coefficients, not the fixed amplitudes required by Hawkes.  In practice, MHP can be implemented event-driven with instantaneous jumps (impulses) plus continuous decay between events.

When using a monotone link $f_k$ (e.g., Softplus) instead of the linear readout, the model becomes a neural Hawkes process generalization. Proposition~\ref{prop:multi-exp} shows the reduction of MHP to classical Hawkes process by selecting a linear, nonnegative readout; with Softplus, one obtains a flexible nonlinearity around the linear regime.
\end{remark}

   \subsection{\textbf{Proposition 2}}
	
	\begin{proof}
		With $A=-\omega I$, 
        \begin{align}
        \overline{\mathbf{A}}(t_i)&=e^{A\Delta_i}=e^{-\omega \Delta_i}I.
        \nonumber \\
        \overline{\mathbf{B}}(t_i) &= (\Delta_i\mathbf{A})^{-1}(\exp(\Delta_i\mathbf{A}) - \mathbf{I}) \cdot \Delta_i\mathbf{B}(t_i)\nonumber\\
        &= \frac{1}{-\omega}(e^{-\omega \Delta_i}-1)\mathbf{B}(t_i) \nonumber\\
        &= \big(1-e^{-\omega \Delta_i}\big)\frac{1}{\omega} \mathbf{B}(t_i). 
        \end{align}
        With $B=\omega I$, $\overline{\mathbf{B}}(t_i)= \big(1-e^{-\omega \Delta_i}\big) I$, giving the claimed recurrence.
	\end{proof}

	\begin{remark}
			RMTPP \cite{du2016recurrent} uses a nonlinear discrete recurrence $h_i=\tanh(W_h h_{i-1}+W_x x_i + W_\Delta \Delta_i + b)$ and a softplus intensity head of the form $\lambda_k(t)=\mathrm{Softplus}(\alpha_k(t-t_i)+w_k^\top h_i + b_k)$. Proposition~\ref{prop:edrnn} shows MHP reduces  to a continuous-time exponentially decayed gated RNN, a well-studied continuous-time RNN variant, not to RMTPP's tanh update. However, the overall intensity form $\lambda_k(t)=f_k(\alpha_k(t-t_i)+w_k^\top h_i + b_k)$ matches the standard RMTPP readout family. Thus, MHP generalizes RMTPP's intensity head while adopting a different, event-driven, selective SSM recurrence with continuous-time consistency and linear-time scanning.
	\end{remark}

\end{document}